\documentclass{article} % For LaTeX2e
\usepackage{iclr2027_conference,times}

\usepackage[utf8]{inputenc} % allow utf-8 input
\usepackage[T1]{fontenc}    % use 8-bit T1 fonts
\usepackage{hyperref}       % hyperlinks
\hypersetup{hidelinks}
\usepackage{url}            % simple URL typesetting
\usepackage{booktabs}       % professional-quality tables
\usepackage{amsmath}
\usepackage{amsfonts}       % blackboard math symbols
\usepackage{amssymb}        % \checkmark, used by the model-set table
\usepackage{nicefrac}       % compact symbols for 1/2, etc.
\usepackage{microtype}      % microtypography
\usepackage{xcolor}         % colors
\usepackage{graphicx}
\usepackage{subcaption}
\usepackage{float}
\usepackage[section]{placeins}
\usepackage{enumitem}

\graphicspath{{}}
\newcommand{\exactFive}{0.922}
\newcommand{\exactFiveTrained}{0.973}
\newcommand{\exactFiveHeldout}{0.700}
\newcommand{\exactTwenty}{0.406}
\newcommand{\exactTwentyTrained}{0.477}
\newcommand{\exactTwentyHeldout}{0.100}

\title{The Planning Limits of Latent World Models}

\author{Ali J. Alrasheed, Basim Azam, Naveed Akhtar \\
The University of Melbourne, Australia
}

\iclrfinalcopy % show authors and remove line numbers
\begin{document}

\maketitle
\lhead{Preprint} % override the ICLR "Published as a conference paper" header

\begin{abstract}
World models offer a promising way to help robots understand how the physical world evolves and plan complex behaviours through imagination. Yet existing studies mainly demonstrate what these models can accomplish, leaving unclear when their predictions remain useful for planning and where they fail. We study this question using action-conditioned predictors built on five frozen self-supervised visual backbones: V-JEPA 2, V-JEPA 2.1, VideoMAEv2, VideoPrism, and DINOv2. We use frozen backbones to test representations intended to transfer across environments. We evaluate these models on diverse Meta-World manipulation tasks and real-robot interactions from BridgeData V2. We find that a world model guides action selection reliably only when the goal lies within, or slightly beyond, the trajectory it imagines during planning. With five-step rollouts, the length the predictor was trained on, the world model ranks actions reliably only for targets five to ten control steps ahead, whereas task goals lie 16 to 53 steps away. Neither an 81-fold larger predictor nor longer-rollout training extends this range; the encoder affects both range and closed-loop success, with V-JEPA 2.1 performing most consistently. More fundamentally, the limit persists under perfect prediction: using the real simulator, success falls from 92\% to 41\% as the target moves from five to twenty steps ahead of a five-step rollout (Figure~\ref{fig:teaser}). Planning therefore requires either longer imagined trajectories or closer subgoals. For distant goals, pure imagination succeeds in 23\% of episodes, planning with feedback (MPC) raises success to 30\%, imagining as far as the goal to 47\%, and nearby expert subgoals to 76\%. Used within its plannable range, a world model can also improve a vision-language-action (VLA) policy: choosing among eight actions the VLA proposes raises its success from 65\% to 77\% across 16 different tasks.
\end{abstract}

\vspace{-2mm}

\section{Introduction}
\vspace{-2mm}
World models offer a promising approach to robot intelligence: a robot can predict how the world may change under different actions and use these imagined outcomes to guide its behaviour. Latent world models perform this prediction in a learned representation space, avoiding reconstruction of every future pixel \citep{lecun2022path,assran2025v}. Combined with self-supervised visual pretraining, they may transfer general knowledge of objects, motion, and physical interaction across environments. However, existing evaluations mainly show what world models can do, such as controlling agents in games and simulated control tasks, planning towards goal images, and picking and placing objects with a real robot arm \citep{hafner2023mastering,hansen2024td,zhou2024dino,assran2025v,hafner2025training}. They say much less about when and why these models fail, so it remains unclear how much of a complex robotic task current models can plan. Prediction accuracy alone cannot answer this question because it does not always correspond to control performance \citep{lambert2020objective,tian2023control,grimm2020value}.

This paper asks two questions: \emph{how far ahead can a latent world model give a reliable planning signal, and what limits or extends this distance?} For the first question, we test whether the model's imagined futures rank expert actions above random alternatives as the target moves further into the future. We call the largest target distance at which this ranking stays reliable the \emph{plannable range}. It depends on the planning setting, such as the number of steps the model imagines. For the second question, we change one factor at a time: the predictor size (up to 81 times larger), the number of steps the predictor is trained to roll out, the number of steps it imagines during planning, and the frozen visual encoder. The five encoders use different pre-training objectives: latent video prediction (V-JEPA 2 and V-JEPA 2.1), pixel reconstruction (VideoMAEv2), video--text contrastive learning (VideoPrism) and image self-distillation (DINOv2). We also replace the world model with the real simulator, which predicts perfectly, to separate prediction errors from the effect of where the target lies. We evaluate on diverse Meta-World manipulation tasks and offline real-robot interactions from BridgeData V2 \citep{walke2023bridgedata}.

\begin{figure}[t]
\centering
\includegraphics[width=\textwidth]{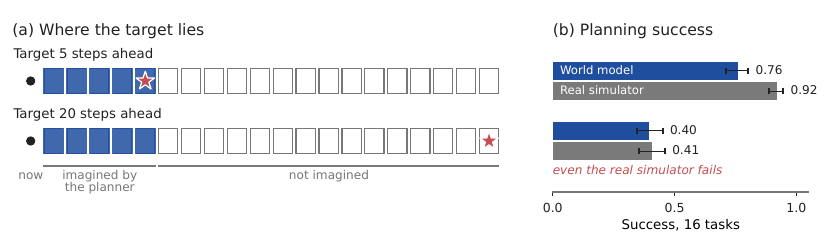}
\caption{\textbf{Planning fails when the target lies beyond the imagined rollout, even with a perfect predictor.} \textbf{(a)} A planner using model predictive control (MPC) imagines five control steps ahead (filled squares) and picks the actions whose imagined states come closest to the target (star), five or twenty steps ahead. \textbf{(b)} Success on sixteen Meta-World tasks with the learned world model (V-JEPA 2.1) and with the real simulator as a perfect predictor.}

\label{fig:teaser}
\end{figure}

Our evaluation leads to the following key findings:
\begin{itemize}[leftmargin=1.5em,itemsep=1pt,topsep=2pt]
\item \textbf{A world model can guide planning only when the target is close to the end of its imagined trajectory.} During planning, the model imagines the states that each candidate action sequence would lead to. When it imagines five control steps ahead, it ranks expert actions above random ones only for targets up to five to ten steps away, but the tasks need 16 to 53 steps to complete. On real-robot data from BridgeData V2, this range is no longer than in simulation (Section~\ref{sec:limits}).
\item \textbf{Where the target lies matters more than predictor size or training.} A larger predictor and training on rollouts longer than five steps do not extend the range. Even the real simulator fails when the target lies beyond its five-step rollout (Figure~\ref{fig:teaser}). Imagining up to the target extends the range to at least twenty steps and improves success (Section~\ref{sec:why}).
\item \textbf{The encoder changes both the range and planning success}, and V-JEPA 2.1 is the most consistent (Section~\ref{sec:encoders}).
\item \textbf{Pure imagination is not enough to complete tasks.} Replanning from new observations helps a little, and nearby subgoals from an expert demonstration help most (Section~\ref{sec:pure}).
\item \textbf{A world model can improve a VLA policy.} Letting the world model choose among the VLA's proposed actions improves the fine-tuned $\pi_0$ policy on the main tasks and on near variants of them (Section~\ref{sec:method}).
\end{itemize}

\vspace{-2mm}
\section{Related work}
\vspace{-2mm}
\paragraph{Latent world models for robot planning.}
World models predict how an environment changes under actions, so an agent can evaluate behaviours before executing them \citep{ha2018world,hafner2019dream,hafner2023mastering,hafner2025training}. Latent world models predict in a learned feature space instead of pixels \citep{lecun2022path,hansen2024td}; I-JEPA, V-JEPA and V-JEPA 2 developed this idea for images, video and action-conditioned prediction \citep{assran2023self,bardes2023v,assran2025v}. Many world models learn their representation from the target environment alone. In this paper, we study world models built on frozen self-supervised encoders, pre-trained on broad visual data to capture general knowledge of the world, with only the action-conditioned predictor trained for the task \citep{zhou2024dino,maes2026leworldmodel,alrasheed2026latent}.
% \vspace{-2mm}
\paragraph{Evaluating world models for planning.}
\vspace{-2mm}
Prediction metrics can disagree with control performance \citep{lambert2020objective,tian2023control,grimm2020value,janner2019trust}, which motivates closed-loop evaluations \citep{zhang2026world,yu2026should}. Because errors compound over long rollouts, earlier work trains on multi-step predictions \citep{talvitie2014model,venkatraman2015improving,hafner2019learning} or keeps rollouts short \citep{janner2019trust}. Other work probes what visual representations encode about manipulation and physics \citep{majumdar2023we,burns2023makes,yeom2026makes,garrido2025intuitive,joseph2026interpreting}. We instead ask how far ahead the imagined outcomes of latent world models built on frozen encoders remain useful for goal-directed action selection, which factors set this distance, and how planning can operate within it.

\vspace{-2mm}
\paragraph{Intermediate goals and action selection.}
Hierarchical planning uses intermediate targets for long tasks \citep{nair2019hierarchical,eysenbach2019search,pertsch2020long}, and recent methods use predictive models to evaluate VLA or generative action proposals before execution \citep{qi2026inference,wu2025foresight,kwok2025robomonkey}. Appendix~\ref{app:related} discusses all three areas in more detail.

\vspace{-1mm}
\section{Evaluation and setup}
\label{sec:setup}
\vspace{-1mm}
\paragraph{Evaluated models.}
Each world model contains a frozen visual encoder and a learned action-conditioned predictor. We evaluate the ViT-L versions of V-JEPA 2 \citep{assran2025v}, V-JEPA 2.1 \citep{mur2026v}, VideoMAEv2 \citep{wang2023videomae}, VideoPrism \citep{zhao2024videoprism}, and DINOv2 \citep{oquab2023dinov2}. None of the encoders is adapted to Meta-World. We use the same predictor architecture for every encoder and tune its training separately for each. Standalone control uses the cross-entropy method to search over candidate action sequences, while VLA experiments pair V-JEPA 2.1 with a fine-tuned $\pi_0$ \citep{black2024pi_0}. Appendix~\ref{app:setup} provides architecture, training and planner details.
\vspace{-1mm}
\paragraph{Datasets and tasks.}
Our main evaluation uses Meta-World \citep{yu2020meta}, which provides a common observation and action space across manipulation tasks involving different objects, motions, and contact requirements. Its success labels, expert demonstrations, and exact state transitions support controlled comparisons between learned prediction and simulator dynamics. The main evaluation contains sixteen tasks: thirteen used for training and three held out. Four articulated-object tasks form the diagnostic set for measuring the aggregate plannable range. We additionally evaluate four near variants and four tasks involving new objects or motion patterns. An offline analysis of 2\,958 WidowX episodes from BridgeData V2 \citep{walke2023bridgedata} examines whether the same behaviour appears in real-robot data. Appendix~\ref{app:setup} lists the tasks and protocols.

\paragraph{Measuring the plannable range.}
At a sampled point in a successful expert demonstration, we compare the next expert action sequence with 100 random alternatives. The world model imagines each sequence for $K$ control steps and compares its predicted trajectory with an expert observation $L$ steps in the future. We call $K$ the \textit{rollout horizon} and $L$ the \textit{target lookahead}; one control step executes two environment actions. Unless stated otherwise, we use $K=5$ and $L\in\{1,2,3,5,10,20\}$.

For an action sequence $\mathbf a$, its score is the closest predicted cosine distance to the target:
\begin{equation}
J_L(\mathbf a)
=
\min_{1\leq k\leq K}
\left[
1-\cos\!\left(
\hat z_{t+k}(\mathbf a),
z^{\mathrm{demo}}_{t+L}
\right)
\right],
\label{eq:score}
\end{equation}
where $\hat z_{t+k}(\mathbf a)$ is the predicted latent state and $z^{\mathrm{demo}}_{t+L}$ is the encoded target observation. Lower scores indicate trajectories that approach the target more closely.

We report the \textit{expert percentile}: the percentage of random sequences that receive a lower score than the expert sequence. Lower is better, chance is 50, and a percentile of 10 means that the expert sequence outranks at least 90\% of the alternatives. We define the \textit{plannable range} $P^*$ as the largest tested $L$ whose task-averaged expert percentile is at most 10. Appendix~\ref{app:pstar} describes sampling, aggregation, threshold sensitivity and non-monotone cases.

The distinction between $K$ and $L$ is important. When $K=5$ and $L=20$, the model imagines five steps towards a target twenty steps away; it does not predict arrival at that target. The plannable range therefore describes a particular model and planning configuration. We fix $K=5$ for the main comparisons and vary it in Section~\ref{sec:why}.

\begin{figure}[t]
\centering
\includegraphics[width=\textwidth]{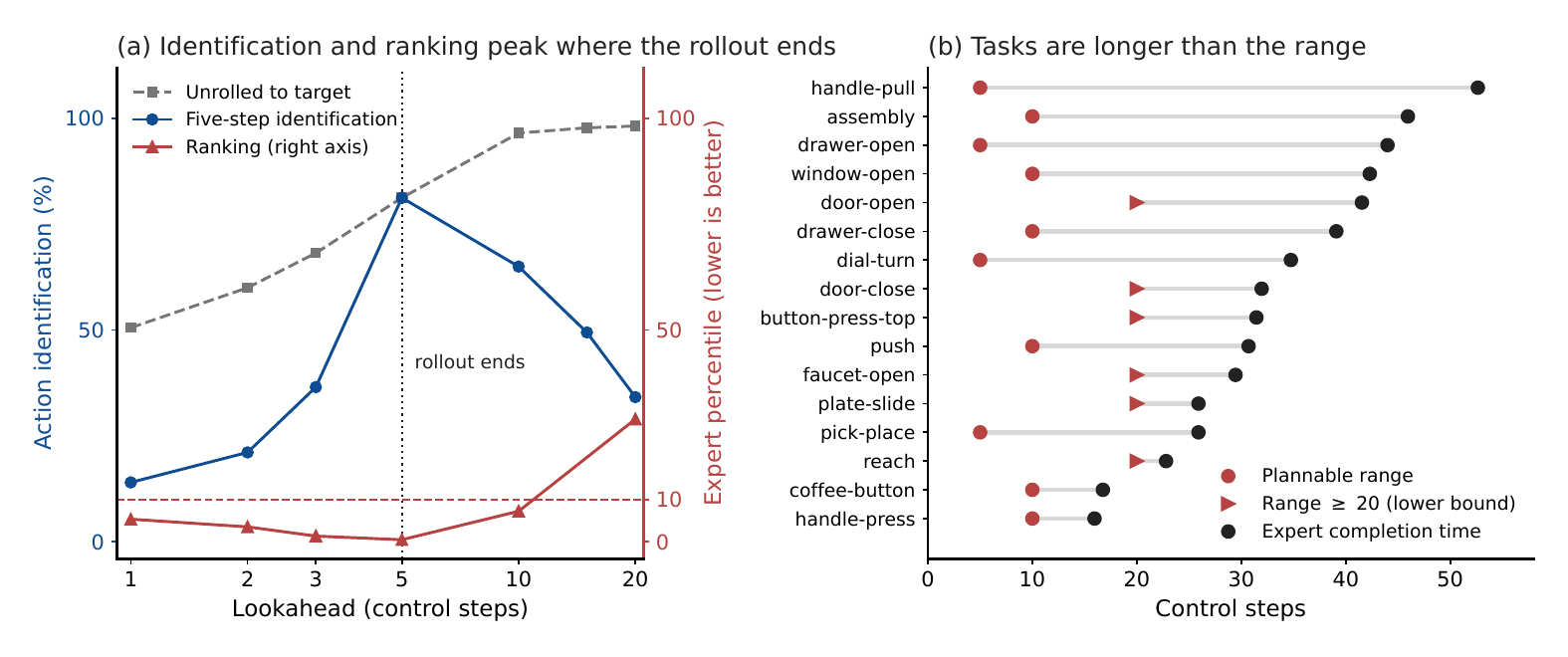}
\caption{\textbf{With a five-step rollout, the world model works best for targets five steps ahead.} \textbf{(a)} V-JEPA 2.1 as the target moves further ahead. Left axis (higher is better): how often the model identifies the executed action sequence among fifteen random ones, imagining five steps (blue) or up to the target (grey). Right axis (lower is better): expert percentile with a five-step rollout, the share of 100 random sequences scored closer to the target than the expert's (red); the dashed line marks the threshold of 10. \textbf{(b)} For each task, the plannable range with a five-step rollout (red) and the control steps the expert needs to finish the task (black). Triangles mark ranges that reach the largest tested lookahead.}
\vspace{-2mm}
\label{fig:range}
\end{figure}

\vspace{-2mm}
\section{How far ahead can world models guide action selection?}
\label{sec:limits}
\vspace{-2mm}

Section~\ref{sec:setup} defined the plannable range as the largest target distance at which imagined outcomes reliably rank expert actions above random ones. We now measure this range. At each point of a successful expert demonstration, we compare the next five expert control steps with 100 random alternatives. We use the demonstration length as a practical measure of task duration: it may not be the shortest solution, but it comes from an expert with a high success rate. The rollout length $K$ is a key setting in model predictive control. It decides how far ahead the planner imagines each candidate, and longer rollouts cost more computation and give prediction errors more steps to build up. World-model planners therefore often use short rollouts \citep{hansen2024td,zhou2024dino}. We use $K=5$, the length our predictors are trained on, for all main comparisons, and vary it in Section~\ref{sec:why} and Appendix~\ref{app:ksweep}.

\subsection{The plannable range is short relative to task completion}

The measured range is ten control steps for V-JEPA 2.1 (Figure~\ref{fig:range}(a)) and between five and ten across the five encoders (Figure~\ref{fig:range_factors}(c)). Expert demonstrations across the sixteen tasks take 16--53 steps, with a typical length of approximately 33 steps; even the strongest measured range therefore covers only about 30\% of a typical demonstration (Figure~\ref{fig:range}(b)). With this rollout, none of the encoders meets the action-ranking threshold when evaluated against the final task observation. Six tasks reach the largest lookahead we tested, 20 steps, so their true range may be longer (triangles in Figure~\ref{fig:range}(b)). Even so, with a five-step rollout, the world model cannot reliably guide the planner towards distant targets. Section~\ref{sec:why} shows that this limit moves when the rollout is extended.

\vspace{-2mm}
\subsection{A similar pattern appears in offline real-robot data}
\vspace{-2mm}
We repeat the ranking test on 2\,958 WidowX episodes from BridgeData V2. As in simulation, the ranking of the recorded actions weakens as the target moves further ahead. The learned predictors have ranges of zero to five dataset steps, no longer than in simulation. The released V-JEPA 2-AC (action-conditioned) model shows the same pattern, although this data differs from its training data: it favours the recorded actions one step ahead but is close to chance ten steps ahead. The robot, action space and control rate differ from Meta-World, so the numbers cannot be compared directly. These results also come from offline data, not from real-robot control. Appendix~\ref{app:setup} gives the protocol and complete curves.

\vspace{-2mm}
\subsection{Prediction and planning share the same rollout dependence}
\vspace{-2mm}
We test whether the world model's predictions stay informative beyond its plannable range by asking it to identify the executed action sequence among fifteen random ones, comparing its prediction for each with the real future. When the model imagines all the way to the target, identification keeps improving as the target moves further ahead (grey curve in Figure~\ref{fig:range}(a)). When action identification and action ranking use the same rollout length, however, they behave alike: both are most reliable when the target lies at the end of the rollout and weaken as it moves nearer or further, for all five encoders (blue and red curves in Figure~\ref{fig:range}(a); Appendix~\ref{app:protocol}). Predictions therefore stay informative beyond the plannable range only when the model imagines far enough to reach the target, so prediction and planning should be compared at the same rollout length.

\vspace{-2mm}
\section{What determines the plannable range?}
\vspace{-2mm}
\label{sec:why}

Section~\ref{sec:limits} measured the plannable range under five-step imagined rollouts. We now test which factors extend it: predictor capacity, recursive training, the length of the imagined rollout, and the frozen encoder. We then examine why the distance-based score fails for distant targets. Complete experimental details are provided in the appendix.

\begin{figure}[t]
\centering
\includegraphics[width=\textwidth]{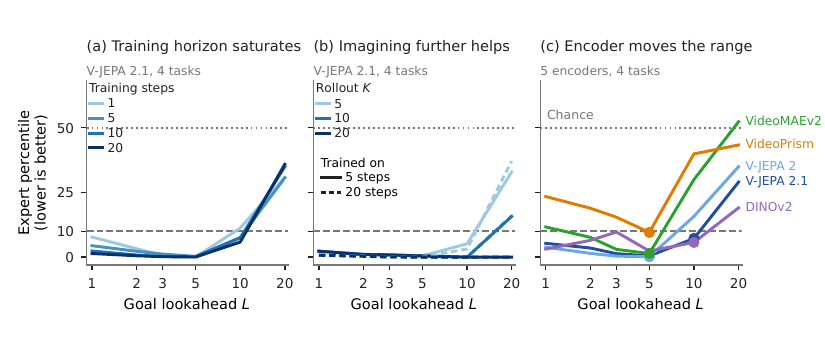}
\caption{\textbf{Only a longer rollout extends the plannable range beyond ten steps.} Expert percentile on the four diagnostic tasks; lower is better, chance is 50 and the threshold is 10. \textbf{(a)} Number of recursive training steps. \textbf{(b)} Rollout length; solid and dashed lines are V-JEPA 2.1 predictors trained with five and twenty recursive steps. \textbf{(c)} Frozen encoder; circles mark each encoder's plannable range.}
\vspace{-10pt}
\label{fig:range_factors}
\end{figure}

\vspace{-2mm}
\subsection{Increasing predictor capacity does not extend the plannable range}
\vspace{-2mm}
Under the five-step planning configuration, increasing predictor capacity by 81 times improves validation loss and latent prediction but leaves the plannable range unchanged. The smallest predictor already reaches the same range, so predictor size is not what limits it in this setting. Appendix~\ref{app:ladder} provides the capacity and prediction results.

\subsection{Recursive training helps, but its benefit saturates quickly}
\vspace{-2mm}
In order to imagine several steps ahead, the predictor works autoregressively: each predicted state becomes the input for the next prediction. A predictor trained for one step only ever sees states encoded from real observations. During planning, however, it receives its own imperfect predictions, so errors can build up. Recursive training addresses this by feeding the predictor's outputs back as inputs over $H_{\mathrm{train}}$ steps. We ask how many training steps are needed and whether more steps extend the range. For V-JEPA 2.1, moving from one-step to five-step training extends the range from five to ten steps, but training on up to twenty steps does not extend it further (Figure~\ref{fig:range_factors}(a)). VideoMAEv2 does not improve at any training horizon. Recursive training therefore helps, but its benefit depends on the representation and stops after a few steps. When the rollout reaches the target, a predictor trained with five recursive steps already ranks expert actions reliably thirty steps ahead, and longer training adds little (Appendix~\ref{app:ladder}).
\vspace{-1mm}
\subsection{Longer imagined rollouts extend the range and improve control}
\label{sec:rollout}
\vspace{-1mm}
In model predictive control (MPC), the planner imagines each candidate action sequence for $K$ steps, scores it by how close the imagined states come to the target, executes the first actions of the best candidate, and then plans again from the new observation. The rollout length $K$ is different from the number of recursive training steps $H_{\mathrm{train}}$: $H_{\mathrm{train}}$ sets how the predictor is trained, whereas $K$ sets how far ahead it imagines during planning. World-model planners usually keep $K$ short, because each extra step costs computation and gives prediction errors more room to build up, and they rely on replanning from new observations to correct the plan \citep{finn2017deep,ebert2018visual,hafner2019learning,hansen2024td,zhou2024dino,assran2025v}. With $K=5$ and a target twenty steps away, however, the imagined states cannot reach the target. When we set $K$ equal to the target distance, V-JEPA 2.1 predictors trained with five and with twenty recursive steps both again rank expert actions reliably up to at least twenty steps, the largest lookahead tested (Figure~\ref{fig:range_factors}(b)). The predictor trained with five recursive steps stays useful even further, despite compounding errors: with a thirty-step rollout, it still ranks the expert reliably against harder alternatives made by perturbing the expert's actions (Appendix~\ref{app:hardneg}). Of the factors tested in this section, only the rollout length extends the range beyond ten steps.

A longer rollout also improves closed-loop control. When planning towards a goal image from the same episode, extending the rollout raises success more than feedback does, even at the same computational budget, and feedback helps most once the rollout reaches the goal (Section~\ref{sec:pure}).

\vspace{-1mm}
\subsection{The encoder changes the range, and V-JEPA 2.1 is the most consistent}
\label{sec:encoders}
\vspace{-1mm}
With the predictor and planner fixed, changing the frozen encoder changes the plannable range between five and ten steps (Figure~\ref{fig:range_factors}(c)). The encoders differ in pre-training objective, data and feature format, so we attribute the effect to the resulting representation as a whole, not to a single pre-training choice. We compare the encoders in three ways: the five-step test above, a matched rollout with harder alternatives made by perturbing the expert's actions, and closed-loop planning. A matched rollout with random alternatives cannot separate them, because all five rank the expert almost perfectly. V-JEPA 2.1 is among the two best encoders in all three comparisons, while the order of the others changes. For example, DINOv2 is best in the five-step test but fourth against harder alternatives (Appendix~\ref{app:hardneg}; Section~\ref{sec:encoders_cl}). The encoder therefore affects the planning signal, and V-JEPA 2.1 is the most consistent.

\subsection{Distance-based scores assume that every step moves closer to the target}
\label{sec:score}

Imagining all the way to a distant target is often too expensive, so planners usually imagine a short rollout and score it by how close its imagined states come to the target, here with cosine distance (Equation~\ref{eq:score}). This score assumes that good actions always move the state closer to the target, and the assumption can fail. Progress may need a temporary detour, such as moving the gripper into position before grasping, which first increases the distance. When the target lies beyond the rollout, good and poor candidates can also end at similar distances from it. Along expert demonstrations, the distance to the target does decrease almost steadily, yet once the target lies beyond the imagined rollout, candidates receive similar scores and become hard to separate (Appendix~\ref{app:weakest}).

This weakness comes from the score, not from the predictor. When we replace the world model with the real simulator, which predicts every outcome exactly, and score candidates by the true distance of the hand and objects to the target, success still falls from 92\% to 41\% as the target moves from five to twenty steps ahead of a five-step rollout (Figure~\ref{fig:teaser}). At twenty steps, the learned model reaches 40\%, close to the simulator. Replacing cosine with Euclidean distance also does not extend the range (Appendix~\ref{app:weakest}). Planning with short rollouts therefore needs either a score that measures progress towards the goal rather than closeness to it, or targets that lie within the rollout, as Section~\ref{sec:pure} shows.

\begin{figure}[t]
\centering
\includegraphics[width=\textwidth]{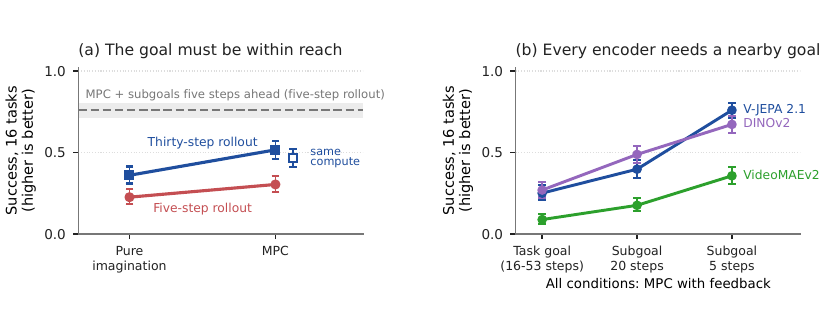}
\vspace{-2mm}
\caption{\textbf{Keeping the target within the imagined rollout improves planning success.} \textbf{(a)} V-JEPA 2.1 on the sixteen main tasks, planning towards a goal image from the same episode with a five-step (red) or thirty-step (blue) rollout, in pure imagination or with MPC. The open square uses the same computation as the five-step planner; the filled thirty-step points use six times more. The dashed line and grey band show MPC with a five-step rollout and subgoals five steps ahead. \textbf{(b)} Three encoders with MPC and a five-step rollout, planning towards a goal image from another reset or towards subgoals twenty or five steps ahead. Subgoals come from an expert rollout of the same episode. }
\vspace{-2mm}
\label{fig:planning}
\end{figure}

\vspace{-2mm}
\section{Is pure imagination enough to complete tasks?}
\label{sec:pure}
\vspace{-2mm}
Sections~\ref{sec:limits} and~\ref{sec:why} tested the world model without letting it control the robot: each test started from a real frame of an expert demonstration and asked only whether the model ranks the expert's next actions above random ones. A robot, however, must choose every action itself until the task is complete, which takes 16 to 53 control steps, while with a five-step rollout the model ranks actions reliably only five to ten steps ahead. This section tests how well the world model plans in closed loop and how much help it needs. We ask three questions. First, can the model complete tasks from pure imagination, planning only from its own predicted states after the first observation? Second, does observing the real outcome after each step help? Third, what helps most: imagining further ahead, which Section~\ref{sec:why} showed extends the range, or giving the planner nearby subgoals? We plan with V-JEPA 2.1 on the sixteen main tasks, with twenty paired episodes per task. The target is an image of the completed task from the same episode, or subgoals from an expert rollout of that episode (Appendix~\ref{app:planner} lists the planner settings). In short, pure imagination rarely succeeds, feedback helps a little, and nearby subgoals help most.

\vspace{-2mm}
\subsection{Pure imagination fails in most episodes}
\vspace{-2mm}
In pure imagination, the planner sees only the first observation. After that, it plans from its own predicted states and never checks the real outcome. With a five-step rollout, it succeeds in 23\% of episodes (Figure~\ref{fig:planning}(a)). Imagining thirty steps ahead raises this to 36\% (58 paired gains against 15 losses), but this uses six times more rollout computation and is still far from reliable. In pure imagination, each new plan starts from a predicted state, so errors can build up over an episode of up to 100 control steps.
\vspace{-2mm}
\subsection{Feedback and replanning help only a little with a short rollout}
\vspace{-2mm}
With model predictive control (MPC), the planner uses the cross-entropy method (CEM) to plan a $K$-step action sequence, executes only its first control step, observes the real outcome, and then plans again from the new observation. It therefore plans once per control step, up to 100 times per episode. With a five-step rollout, this raises success only from 23\% to 30\% (Figure~\ref{fig:planning}(a)). Feedback corrects the current state, but the planner still compares a short imagined trajectory with a distant target, the weakness described in Section~\ref{sec:score}. Feedback helps more when the rollout is long enough to reach the goal: with a thirty-step rollout, it raises success from 36\% to 52\% (70 gains against 20 losses). An earlier comparison over twelve encoder--task conditions shows the same small effect of feedback with a five-step rollout (Figure~\ref{fig:regimes} in Appendix~\ref{app:blind}).
\vspace{-2mm}
\subsection{Nearby subgoals help most}
\label{sec:encoders_cl}
\vspace{-2mm}
Keeping MPC and the five-step rollout fixed, we replace the final goal image with subgoals five control steps ahead, taken from an expert rollout of the same episode. Success rises from 30\% to 76\% (146 paired gains, no losses; Figure~\ref{fig:planning}(a)). This is well above the 47\% that a thirty-step rollout towards the goal image reaches at the same budget (open square in Figure~\ref{fig:planning}(a)). The gain needs the learned dynamics, not only the subgoals: in an earlier comparison over twelve encoder--task conditions, the same planner with a randomly initialised predictor and the same subgoals succeeds in only 0.8\% of episodes (Appendix~\ref{app:blind}). A world model that is reliable over $K$ steps can therefore guide a longer task when each target lies within $K$ steps. Appendix~\ref{app:profile} describes subgoal selection, and Figure~\ref{fig:blind} in Appendix~\ref{app:blind} examines how long the controller can act between observations. 

The benefit of nearby subgoals holds across encoders. For all three encoders tested in closed loop, success rises as the target moves closer, from a goal image of another reset to subgoals twenty and then five steps ahead (Figure~\ref{fig:planning}(b)). The best encoder changes with the target: DINOv2 is best for the two distant targets and V-JEPA 2.1 for five-step subgoals, while VideoMAEv2 is lowest in all three, consistent with Section~\ref{sec:encoders}. These subgoals, however, come from an expert rollout of the same episode, and obtaining them without one remains an open problem.

% \vspace{-2mm}
% \subsection{Encoder choice changes success, but nearby subgoals help all three}
% \label{sec:encoders_cl}
% \vspace{-2mm}

% Figure~\ref{fig:planning}(b) repeats the comparison for three encoders, planning towards a goal image from another reset or towards subgoals twenty or five steps ahead. The best encoder depends on the target: DINOv2 is best with the goal image and the twenty-step subgoals, V-JEPA 2.1 with the five-step subgoals, and VideoMAEv2 is lowest in all three, matching the changing order in Section~\ref{sec:encoders}. Every encoder improves strongly as the subgoals move closer, so the benefit of nearby subgoals does not depend on one representation. However, these subgoals come from an expert rollout of the same episode, and obtaining them without one remains an open problem.

\begin{figure}[t]
\centering
\includegraphics[width=\textwidth]{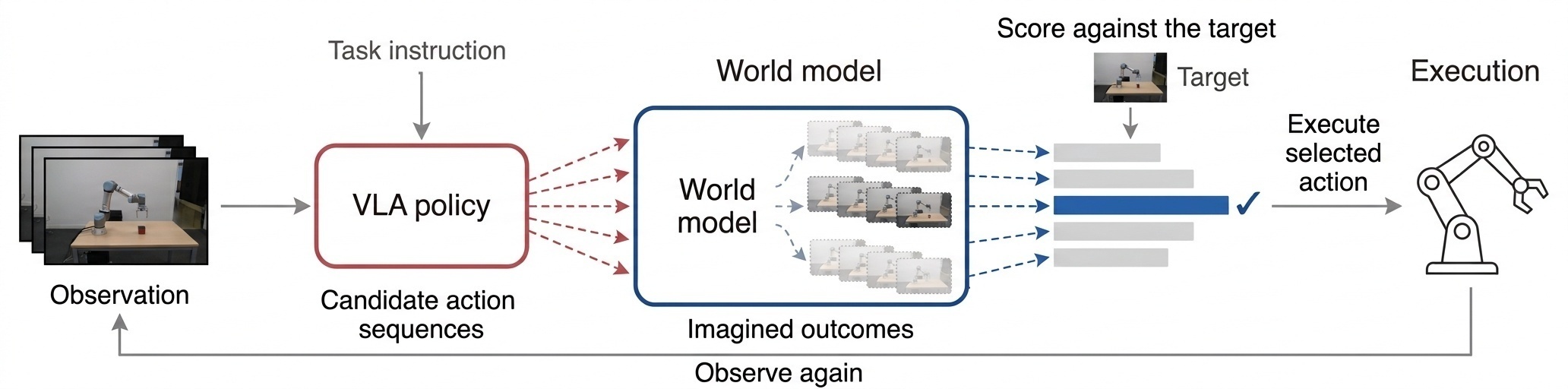}
\caption{\textbf{VLA--WM control.} The VLA proposes $N$ candidate action sequences from the current observation and the task instruction. The world model imagines and scores each candidate against the target, and the robot executes the best one and then observes again.}
\vspace{-2mm}
\label{fig:schematic}
\end{figure}

\vspace{-2mm}
\section{Can world models improve VLA? Towards combined control}
\label{sec:method}
\vspace{-2mm}
Vision-language-action (VLA) models have become a mature and rapidly advancing approach to robot control: they map images and a language instruction directly to actions \citep{black2024pi_0,kim2024openvla}. We ask whether a world model can make such a policy better. The two are complementary. A VLA proposes plausible, task-conditioned actions, but it does not check what those actions will lead to. A world model can predict and compare the outcomes of actions, but on its own it must search a large action space, most of which is irrelevant to the task. Combining them lets each cover the other's weakness: the VLA narrows the search to a few plausible candidates, and the world model checks them before execution. At each control step, the VLA samples $N$ candidate action sequences, and the world model imagines the outcome of each for $K$ control steps. The robot then executes the first control step of the candidate whose imagined states come closest to the target and observes again (Figure~\ref{fig:schematic}). We test this with a fine-tuned $\pi_0$ policy and V-JEPA 2.1, using $N=8$. This tests whether world-model selection can improve a policy when it is used within its plannable range. It is not yet a controller that works without demonstrations. Appendix~\ref{app:profile} gives the fine-tuning, subgoal selection and execution details.

\begin{figure}[!htbp]
\centering
\includegraphics[width=\textwidth]{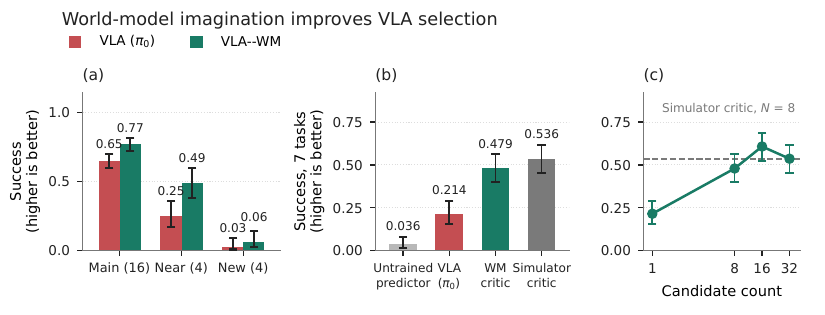}
\vspace{-3mm}
\caption{\textbf{World-model selection improves the VLA on trained tasks and near variants.} \textbf{(a)} Success on the main and held-out task groups. \textbf{(b)} Learned, untrained and simulator-based critics on a seven-task subset. \textbf{(c)} Candidate-count sweep on the same subset.}
\vspace{-4mm}
\label{fig:method}
\end{figure}
\vspace{-2mm}
\subsection{A world model can improve VLA action selection}
\vspace{-2mm}
With $N=8$ proposals and subgoals, world-model selection raises success from 65\% to 77\% across the sixteen main tasks, and no task becomes worse (Figure~\ref{fig:method}(a)). The gain needs the learned dynamics: an untrained predictor given the same subgoals, score and candidates performs below the VLA alone (Figure~\ref{fig:method}(b)). The learned world model is also close to an ideal selector. On the seven tasks with room for improvement, it achieves 82\% of the gain obtained by scoring candidates with the simulator. Selection also helps on held-out variants of trained object families, raising success from 25\% to 49\%. On tasks with new objects or motions, however, both methods stay near zero (Table~\ref{tab:heldout_pairing} in Appendix~\ref{app:profile}). The world model therefore improves the policy when the policy already proposes some useful actions, but it cannot fix a policy whose proposals are all poor.

\vspace{-3mm}
\subsection{VLA proposals reduce the candidate search}
\vspace{-2mm}
A world model can imagine the outcome of an action sequence, but it cannot propose actions itself. Used alone, it relies on a planning optimiser such as CEM, which samples hundreds or thousands of candidate action sequences, scores each with the world model, and refines the sampling distribution over several iterations. Most of these candidates are irrelevant to the task. Our standalone planner, for example, scores hundreds of candidates at every control step. With VLA proposals, the world model scores only $N$ task-conditioned candidates. A small $N$ is enough: on the seven-task subset, success peaks at sixteen candidates and stops improving beyond that (Figure~\ref{fig:method}(c)). This is a count of candidates, not of computation time, which also depends on the size of the policy. The policy and the world model therefore play separate roles: the policy decides which actions are worth considering, and the world model decides which of them to execute.
\raggedbottom
\vspace{-2mm}
\section{Conclusion}
\vspace{-2mm}
We asked how far ahead latent world models can guide the choice of robot actions, and what limits this distance. A world model guides action selection reliably only when the target lies within, or slightly beyond, the trajectory it imagines. With a five-step rollout, this is five to ten control steps, while the tasks take 16 to 53. An 81-fold larger predictor and training on longer rollouts do not extend this range; the encoder changes it only between five and ten steps, and V-JEPA 2.1 is the most consistent of the five. The range is limited mainly by where the target lies: even a planner that uses the real simulator falls from 92\% to 41\% success when the target moves from five to twenty steps ahead of a five-step rollout, so imagining up to the target, rather than better prediction alone, extends the range and improves control.

These results suggest three lessons for planning with latent world models. Evaluate how far ahead a model can rank actions, not only how well it predicts: the larger predictor predicted better but did not plan further. Keep the target within the imagined trajectory: a rollout long enough to reach the goal raised success from 30\% to 47\% at the same computational cost. Combine feedback with such a target: on its own, feedback raised success only from 23\% to 30\% with a five-step rollout, but from 36\% to 52\% with a thirty-step rollout. As a first step towards combined control, the world model also improved a VLA policy from 65\% to 77\% by selecting among its proposed actions. The main limitation of current latent world models is that they complete long tasks well only when given targets close enough to lie within their imagined trajectory, which are not always available, so future world models will need to generate their own intermediate targets or plan reliably towards distant goals (Appendix~\ref{app:limits}).

\bibliographystyle{iclr2027_conference}
\bibliography{iclr2027_conference}

\appendix
\clearpage
\section*{Appendix}

\section{Extended related work}
\label{app:related}
Model-based reinforcement learning often learns the representation, dynamics and policy together from interactions with the target environment \citep{hafner2019dream,hafner2023mastering,hafner2025training}. Earlier work limits compounding errors with multi-step training or short rollouts \citep{talvitie2014model,janner2019trust}; we find that extending the rollout until it reaches the target still improves action ranking and control, although the gain is smaller than with exact dynamics. Our plannable-range analysis separates predictive ability from goal-directed action selection and examines how the range changes with the visual representation, predictor design, planning objective, rollout horizon and search procedure.

The distinction between prediction and decision quality connects this study to objective mismatch in model-based control \citep{lambert2020objective,grimm2020value}. A predictor can preserve information about observed transitions without supplying a useful ordering of actions under a chosen goal-distance objective. We measure both action identification and goal-directed ranking, and Section~\ref{sec:limits} shows that they share the same dependence on the rollout.

Hierarchical planning addresses distant goals by introducing intermediate targets \citep{nair2019hierarchical,eysenbach2019search,pertsch2020long}. Our experiments take intermediate targets from expert demonstrations. This isolates whether a world model can use a nearby target, and it is complementary to methods that generate such targets.

Vision-language-action models such as $\pi_0$ and OpenVLA generate robot actions from visual observations and language instructions \citep{black2024pi_0,kim2024openvla}. GPC uses predictive world modelling to evaluate generative action proposals \citep{qi2026inference}; FOREWARN combines latent outcome prediction with language-based evaluation \citep{wu2025foresight}; and RoboMonkey studies inference-time improvement through verification \citep{kwok2025robomonkey}. Our experiments use a fine-tuned $\pi_0$ to generate task-conditioned proposals and a latent world model to evaluate their predicted outcomes. We compare learned, untrained, and simulator-based evaluation and study how performance changes across candidate counts and held-out tasks. Appendix~\ref{app:profile} specifies the information each system receives.

\subsection{Relation to other world-model families}
These results describe action-conditioned predictors trained on frozen pretrained encoders. The encoder is not updated for control, and the predictor learns dynamics in a representation chosen for its generality rather than for the task. This is the setting of the released video models we evaluate, and it is the reason the encoder can be varied while everything else is held fixed.

Other families learn the representation together with the planning objective. TD-MPC2 learns its latent jointly with the objective it is planned against \citep{hansen2024td}, value-equivalent models learn a representation that preserves returns rather than futures \citep{grimm2020value}, and DINO-WM keeps a frozen encoder but uses a different goal-distance construction \citep{zhou2024dino}. Testing whether the same relation between rollout length and target distance holds for these families is a natural next step. We expect the methodological lesson to apply more widely: a short measured planning range should be tested by varying the rollout before it is attributed to the representation.

\section{Models, tasks and evaluation metrics}
\label{app:setup}
\subsection{World-model architecture and training}
The five frozen encoders are the ViT-L releases of V-JEPA 2, V-JEPA 2.1, VideoMAEv2, VideoPrism, and DINOv2 cited in Section~\ref{sec:setup}. Encoder weights are not adapted on Meta-World. The feature extractor resamples a common sixteen-environment-step observation history to each encoder's input format. Images are rendered at $256\times256$ from the simulator's corner camera. Training features are standardised using statistics from the training split.

Each baseline dynamics predictor has 3\,754\,240 parameters, transformer width 256, four blocks, four attention heads, an MLP expansion ratio of four, and a four-state history. It projects visual tokens and two-action blocks into a common hidden space, uses causal attention over time, and predicts a residual update to the last latent state. Spatial token positions use fixed sinusoidal encodings, avoiding a learned parameter count that changes with token count. The main closed-loop checkpoints use a $2\times4\times4$ token grid for V-JEPA 2, V-JEPA 2.1, VideoPrism, and DINOv2, and $2\times7\times7$ for VideoMAEv2. The four-task diagnostic evaluation uses pooled tokens. Equal predictor parameter counts therefore do not imply equal token counts across settings.

Predictors are trained autoregressively over five steps by default. The loss averages mean-squared latent error and cosine distance to the target latents, with equal weights, across rollout steps. AdamW uses learning rate $10^{-3}$, weight decay 0.05, and momentum parameters $(0.9,0.999)$. The effective batch size is 64, accumulated from microbatches of sixteen. Training uses a five-percent warm-up and cosine learning-rate decay to $10^{-5}$, with a forty-epoch limit and early-stopping patience of three validation epochs. The lowest-validation-loss checkpoint is selected. The baseline configuration and optimiser are shared across encoders; capacity and horizon comparisons explicitly vary the relevant settings.

Episodes are split before constructing windows. Every fifth episode in each task's sorted cache is held out for validation, producing an approximately 80/20 split across expert and random-action collection modes. No window crosses the training-validation boundary. Task-level held-out evaluations are separate from this within-task validation split. The reported checkpoints use training seed zero, and additional seeds are used for selected range checks (Table~\ref{tab:dispersion} and Figure~\ref{fig:real}).

\subsection{Task sets and observation access}
The thirteen training tasks are assembly, button-press-topdown, coffee-button, dial-turn, door-close, door-open, drawer-close, drawer-open, faucet-open, handle-press, pick-place, plate-slide, and reach. The three held-out tasks in the main sixteen-task evaluation are window-open, handle-pull, and push. All names refer to Meta-World v3 environments. The four-task diagnostic set contains door-open, door-close, drawer-open, and drawer-close.

The additional near variants are faucet-close, plate-slide-side, handle-press-side, and button-press. The additional new tasks are soccer, lever-pull, hammer, and basketball. These eight tasks are excluded from predictor and VLA fine-tuning and are reported as two groups of four. A scripted-expert gate uses a success threshold of 95\% before including a task. Each gate uses ten episodes and a 200-environment-step limit. Expert completion times therefore describe a representative, high-success expert rather than the shortest possible solution.

The adapted VLA receives the current RGB image, a task-specific language instruction, and a seven-dimensional state input containing gripper position, gripper opening, and object position, where object position is read from the simulator. World-model planning receives either demonstration subgoals or a single goal image, as stated with each result; VLA--WM selection receives demonstration subgoals. Appendix~\ref{app:profile} describes how each system executes its actions.

\subsection{Planner settings}
\label{app:planner}
All closed-loop experiments with the learned world model replan at every control step. Apart from VLA--WM selection, which ranks VLA proposals instead (Appendix~\ref{app:profile}), they use the same planner, the cross-entropy method (CEM). Table~\ref{tab:planner} lists the settings that change between experiments. Each candidate is a sequence of $K$ action blocks, where one block is one control step of two four-dimensional environment actions. In the first iteration, actions are sampled from a Gaussian with standard deviation one and clipped to $[-1,1]$. Its mean is zero at the first step of an episode. At later steps, the mean is the previous solution shifted forward by one block, with zeros for the new last block (warm start). The predictor starts from a history of four latent states and imagines each candidate for $K$ control steps. At the start of an episode, this history repeats the first observation with zero actions. Each candidate is scored with Equation~\ref{eq:score}, the smallest cosine distance to the target along the imagined trajectory. The mean and standard deviation of the elites, the candidates with the lowest scores, define the sampling distribution of the next iteration. After three iterations, the robot executes the first block of the final mean. With MPC, the new observation is encoded and added to the history before the next plan. With pure imagination, only the first observation is encoded, and the predicted state is added instead. A subgoal is chosen by finding the demonstration state closest to the current latent state in cosine similarity and moving $L$ control steps further along the demonstration. An episode ends at task success or after 200 environment actions, which is 100 control steps. The search-budget comparison in Appendix~\ref{app:ladder} multiplies the numbers of candidates and elites by ten.

\begin{table}[htbp]
\centering
\caption{Planner settings in the closed-loop experiments. $K$ is the rollout horizon in control steps. Every CEM planner uses three iterations and a warm start. ``MPC'' encodes a new observation at every control step, and ``none'' encodes only the first. A subgoal lies $L$ control steps ahead on an expert rollout of the same episode. A final frame is the last frame of an expert rollout, from the same episode or from another reset, as stated with each result.}
\label{tab:planner}
\small
\setlength{\tabcolsep}{4pt}
\begin{tabular}{llrrll}
\toprule
Experiment & Reported in & $K$ & Cand.\,/\,elites & Feedback & Target \\
\midrule
Goal image, 16 tasks & Fig.~\ref{fig:planning}(a), Tab.~\ref{tab:x30} & 5 & 100 / 10 & MPC or none & final frame \\
 & & 30 & 100 / 10 & MPC or none & final frame \\
 & & 30 & 17 / 2 & MPC & final frame \\
Encoders, subgoals & Figs.~\ref{fig:planning}, \ref{fig:profile} & 5 & 100 / 10 & MPC & $L=5$, 20 or final frame \\
Far subgoals, 8 tasks & Tab.~\ref{tab:closedloopK} & 5 & 100 / 10 & MPC & $L=20$ \\
 & & 20 & 100 / 10 & MPC & $L=20$ \\
 & & 20 & 25 / 3 & MPC & $L=20$ \\
Twelve conditions & Fig.~\ref{fig:regimes} & 5 & 100 / 10 & MPC or none & $L=5$ or final frame \\
Observation interval & Fig.~\ref{fig:blind} & 5 & 100 / 10 & every 1--25 steps & $L=5$ \\
VLA--WM selection & Sec.~\ref{sec:method} & 5 & 8 proposals & MPC & $L=5$ \\
\bottomrule
\end{tabular}
\end{table}

\subsection{BridgeData evaluation}
The BridgeData V2 adaptation contains 2\,958 usable WidowX episodes. Its seven-dimensional state contains end-effector position, orientation, and gripper state, with no object position. Its seven-dimensional actions also differ from Meta-World's four-dimensional actions. The dataset has no success labels; demonstrations are retained without a measured success filter. Offline scores on this corpus are therefore not directly comparable with success-gated simulator scores.

\begin{figure}[htbp]
\centering
\includegraphics[width=\textwidth]{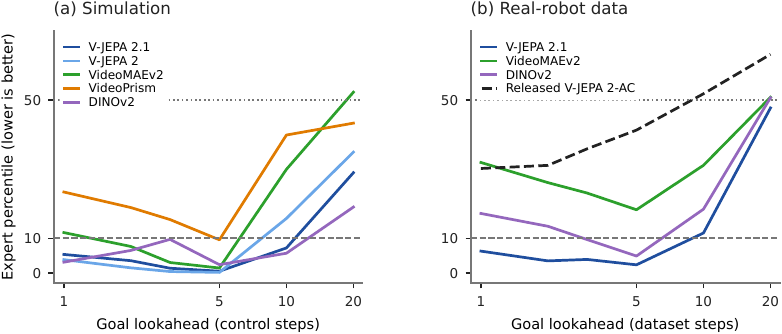}
\caption{Expert-action percentile in Meta-World simulation and WidowX recordings. Lower is better; dashed and dotted lines mark the threshold and chance. Candidate sampling and control rates differ across datasets, and V-JEPA 2-AC is evaluated outside its post-training distribution.}
\label{fig:real}
\end{figure}

A Bridge step is one recorded frame at 5\,Hz, whereas a Meta-World control step groups two environment actions. We report lookahead in the corresponding dataset steps and do not equate physical time across the datasets. Short episodes cannot contribute goals beyond their remaining length. Corpus-matched Gaussian candidate actions use the recorded per-dimension action scale rather than a simulator-sized uniform distribution. The learned-predictor curves average twenty-episode sweeps across three seeds. Figure~\ref{fig:real} reports these offline sweeps.

The released V-JEPA 2-AC control uses its published ViT-g encoder and action-conditioned predictor without retraining. The evaluation samples 100 episodes, uses five-step rollouts and 100 random alternatives, and disables extrinsic-camera conditioning. Actions are obtained from recorded state differences with the model-specific preprocessing. This corpus differs from the model's action-conditioned training data, so the result shows how the released model transfers to a new robot dataset.

\section{The plannable range: estimator and protocol}
\label{app:pstar}
\subsection{Goal-distance ranking}
For each successful expert episode, the simulator-based sweep selects starting points at fractions 0.15, 0.35, 0.50, 0.65, and 0.85 of the episode. Starting points lacking sufficient history or future actions are skipped. The four-state latent history and action blocks initialise the predictor. The expert's next five action blocks and 100 random five-block sequences are rolled out. The standard simulator candidate distribution is a standard Gaussian clipped to $[-1,1]$ in each action dimension. Every block contains two four-dimensional environment actions.

For goal lookahead $L\in\{1,2,3,5,10,20\}$, the reference is the expert observation $L$ control steps after the starting point. Equation~\ref{eq:score} takes the minimum goal distance across the five predicted steps. The expert percentile at one window is
\begin{equation}
q_L=\frac{100}{M}\sum_{i=1}^{M}\mathbf{1}\{J_L(\mathbf a_i)<J_L(\mathbf a_{\mathrm{expert}})\},\qquad M=100.
\end{equation}
Ties are not counted as beating the expert, so chance is 50 for continuous scores. The percentile is averaged over valid windows within each task; aggregate encoder curves then give each contributing task equal weight. The per-task analysis retains each task's own curve.

The reported $P^*$ is the largest tested lookahead with mean percentile at most 10, or zero if none passes. A zero is failure of this diagnostic criterion, not zero achievable success. If the maximum tested lookahead passes, the range is right-censored. Non-monotone curves can pass after failing at a shorter lookahead, so the scalar does not imply that every smaller lookahead passes. This is why the curves accompany the scalar.

\subsection{Window populations and threshold sensitivity}
Goals beyond the end of an episode are omitted, rather than replaced with its final frame. Consequently, the number and composition of valid windows can change with $L$. Cached-feature probes also differ from simulator probes: cached collection can include post-success frames and noisy expert actions, whereas the simulator probe stops at success. The action-identification and goal-ranking curves therefore use different window sets.

Table~\ref{tab:thresholds} recomputes ranges from the same aggregate curves at three reporting thresholds. At every threshold, all five-step ranges remain at or below twenty control steps, shorter than the typical task length of about 33 steps, so the main conclusion does not depend on this choice. The encoder comparison in Appendix~\ref{app:hardneg} uses identical windows for all five encoders, at 49, 40, and 60 per task, so no encoder can gain or lose windows by being easier or harder to score.

\begin{table}[htbp]
\centering
\caption{Plannable range in control steps at three percentile thresholds, computed from the same four-task aggregate curves with a five-step rollout. A value of twenty reaches the largest tested lookahead. At every threshold, the ranges remain shorter than typical task lengths.}
\label{tab:thresholds}
\begin{tabular}{lrrr}
\toprule
Encoder & Threshold 5 & Threshold 10 & Threshold 20 \\
\midrule
DINOv2 & 5 & 10 & 20 \\
VideoMAEv2 & 5 & 5 & 5 \\
VideoPrism & 0 & 5 & 5 \\
V-JEPA 2.1 & 5 & 10 & 10 \\
V-JEPA 2 & 5 & 5 & 10 \\
\bottomrule
\end{tabular}

\end{table}

\subsection{Prediction error and recorded-action identification}
Prediction error is evaluated on validation episodes using a ridge decoder fitted on training episodes only. The decoder uses up to 8000 training rows, a PCA projection to at most 512 dimensions, and ridge regularisation of 1.0. PCA and feature standardisation are fitted only on those training rows. The primary Meta-World error is mean Euclidean object-position error in millimetres, evaluated on transitions with more than 5\,mm of object displacement. No-motion and constant-velocity baselines start from decoded current and previous states, so they include decoder error; the mean baseline predicts the training-state mean. We report error curves by lookahead rather than a scalar prediction horizon, because a scalar based on consecutive baseline wins is sensitive to a single short-horizon exception.

Recorded-action identification uses 720 validation windows from the four articulated tasks. For each window, the model predicts the future under the recorded action sequence and fifteen uniformly sampled alternatives in $[-1,1]$. At each prediction horizon, mean-squared latent distance compares each predicted endpoint with the observed future endpoint. The test succeeds if the recorded sequence ranks first; chance is $1/16$. This reference future is only available offline. Separate perturbation tests replace uniform alternatives with noise around the recorded actions. Appendix~\ref{app:hardneg} reports them at a matched rollout, where they resolve differences between encoders that the random-action test cannot.

\begin{figure}[htbp]
\centering
\includegraphics[width=0.75\textwidth]{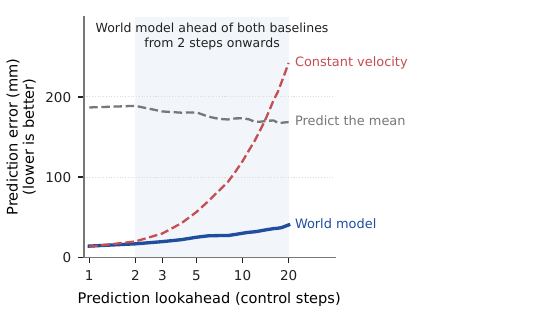}
\caption{Decoded object-position error versus prediction lookahead for V-JEPA 2.1 and two baselines, constant velocity and predict-the-mean. Lower is better.}
\label{fig:prediction}
\end{figure}

\section{The rollout horizon}
\label{app:khorizon}
Equation~\ref{eq:score} contains two horizons. The rollout horizon $K$ sets how many steps the predictor imagines for each candidate; the goal lookahead $L$ selects the demonstration frame those imagined states are compared against. Unless stated otherwise, the range measurements use $K=5$. This appendix shows how that choice shapes the measurement and then sweeps $K$ directly.

Candidate scores are most widely separated when the goal sits at the end of the rollout. Table~\ref{tab:dispersion} reports the relative spread of the goal-distance score across the random candidates at one window, $(\bar c - c_{\min})/\bar c$, where $\bar c$ is the mean and $c_{\min}$ the minimum over the candidate set. The spread is reported relative because absolute cosine distance grows about twentyfold from $L=1$ to $L=20$ as the goal moves away, so an absolute spread would rise while candidates became harder to separate. For all five encoders the spread is largest at $L=5$, where $L$ equals $K$, and falls on both sides of it.

The changing window population does not account for this. The lookahead curve is measured on a moving set of windows: 394 are available at $L\leq5$, 320 at $L=10$, and 212 at $L=20$. The last row of Table~\ref{tab:dispersion} instead uses the fixed-window condition, which retains 36 windows at every lookahead on one task. There the spread falls from 0.403 at $L=5$ to 0.034 at $L=20$, a factor of twelve, and the expert-to-random contrast falls from 0.954 to 0.070 over the same interval.

Both the scoring configuration and the representation remain in play at the longest lookahead. Score concentration at $L>K$ reduces what any ranking can resolve. The window-weighted expert percentile at $L=20$ nevertheless still separates encoders, from 21.1 for DINOv2 to 52.2 for VideoMAEv2, so at $K=5$ the measurement is not saturated. At a matched $K=20$ it is: against random candidates the expert ranks first in every scored window for three of the five encoders, and in almost every window for the other two, so that comparison cannot order them. Appendix~\ref{app:hardneg} resolves the comparison with harder alternatives. Separating the two contributions requires varying $K$ jointly with $L$, which the sweep below does.

\begin{table}[htbp]
\centering
\caption{Relative spread of the goal-distance score across random candidates, by goal lookahead, at a fixed five-step rollout. The largest value in each row is bold and falls at $L=5$ for every encoder. The three DINOv2 seeds agree to within 0.02 at every lookahead; seed 0 is shown. The final row is the fixed-window condition on one task, which holds the window population at 36 across the whole curve.}
\label{tab:dispersion}
\begin{tabular}{lrrrrrr}
\toprule
Encoder & $L=1$ & $L=2$ & $L=3$ & $L=5$ & $L=10$ & $L=20$ \\
\midrule
DINOv2 & 0.043 & 0.044 & 0.061 & \textbf{0.173} & 0.102 & 0.047 \\
VideoMAEv2 & 0.130 & 0.149 & 0.179 & \textbf{0.259} & 0.164 & 0.071 \\
VideoPrism & 0.016 & 0.021 & 0.027 & \textbf{0.059} & 0.052 & 0.039 \\
V-JEPA 2.1 & 0.071 & 0.142 & 0.243 & \textbf{0.401} & 0.168 & 0.066 \\
V-JEPA 2 & 0.043 & 0.082 & 0.133 & \textbf{0.230} & 0.151 & 0.078 \\
V-JEPA 2.1, fixed window & 0.085 & 0.106 & 0.201 & \textbf{0.403} & 0.104 & 0.034 \\
\bottomrule
\end{tabular}

\end{table}

\subsection{Sweeping the rollout horizon}
\label{app:ksweep}
We vary $K$ directly. Two predictors, trained with five and twenty recursive steps, are evaluated at $K\in\{5,10,20\}$ on four articulated tasks, with the goal lookahead swept as before. Table~\ref{tab:kcurve} reports the mean expert percentile, averaged over tasks without weighting.

Setting the rollout horizon equal to the goal lookahead removes the decay. At $K=5$ the percentile at $L=20$ is 32.90 for the five-step predictor and 37.07 for the twenty-step predictor. At $K=20$ both read 0.00, meaning the expert sequence ranks ahead of all 100 random candidates in every scored window of every task. $P^*$ moves from 10 to 20 for both predictors. Twenty is the largest lookahead in this sweep, so $P^*=20$ is a lower bound rather than a measured value. The gain is not linear in $K$: at $K=10$ the same quantity reads 15.71 and 15.86, above the threshold of 10.

On one task the $K=5$ score is not only weak at $L=20$ but reversed. On drawer-close the expert ranks at percentile 74.55 for the five-step predictor and 94.92 for the twenty-step predictor, both worse than chance, and at 0.00 for both at $K=20$. Expert-to-random contrast on that task rises from 0.114 to 0.645.

Window populations are matched across $K$. A probe at block index $p$ in an episode of $n$ blocks is scored only if $p+K<n$, so a window at the boundary $p+20=n$ is scored at $K=5$ and dropped at $K=20$. Twenty-seven windows across the sweep differ in this way, and the episode-length rule drops none. The per-$K$ window sets are nested, so we restrict every condition to the windows common to all three values of $K$. The unrestricted populations give 30.90 and 35.91 at $K=5$ against 0.00 at $K=20$, and the same conclusion.

Appendix~\ref{app:closedloopK} tests the same change in closed-loop control, holding either the search or the total budget fixed.

\begin{table}[htbp]
\centering
\caption{Mean expert percentile by rollout horizon $K$ and goal lookahead $L$, on four articulated tasks with matched window populations. Lower is better; chance is 50 and the reporting threshold is 10. $P^*$ is the largest tested $L$ at or below the threshold. The $L=20$ column at $K=20$ and the resulting $P^*$ are bold. Twenty is the largest lookahead in this sweep, so $P^*=20$ is a lower bound.}
\label{tab:kcurve}
\begin{tabular}{llrrrrrrr}
\toprule
Predictor & $K$ & $L=1$ & $L=2$ & $L=3$ & $L=5$ & $L=10$ & $L=20$ & $P^*$ \\
\midrule
Five-step & 5 & 2.27 & 1.13 & 0.77 & 0.51 & 5.16 & 32.90 & 10 \\
 & 10 & 2.26 & 1.06 & 0.92 & 0.45 & 0.00 & 15.71 & 10 \\
 & 20 & 2.33 & 1.00 & 0.84 & 0.47 & 0.00 & \textbf{0.00} & \textbf{20} \\
\midrule
Twenty-step & 5 & 0.77 & 0.44 & 0.05 & 0.07 & 3.16 & 37.07 & 10 \\
 & 10 & 0.80 & 0.42 & 0.12 & 0.04 & 0.00 & 15.86 & 10 \\
 & 20 & 0.75 & 0.33 & 0.06 & 0.04 & 0.00 & \textbf{0.00} & \textbf{20} \\
\bottomrule
\end{tabular}

\end{table}

\section{Matching the rollout in closed-loop control}
\label{app:closedloopK}
Matching the rollout to the goal improves closed-loop control with the learned predictor, and not only under exact dynamics. Appendix~\ref{app:ksweep} showed that setting $K=L$ restores offline ranking. This section makes the same change in closed loop.

\begin{figure}[htbp]
\centering
\includegraphics[width=\textwidth]{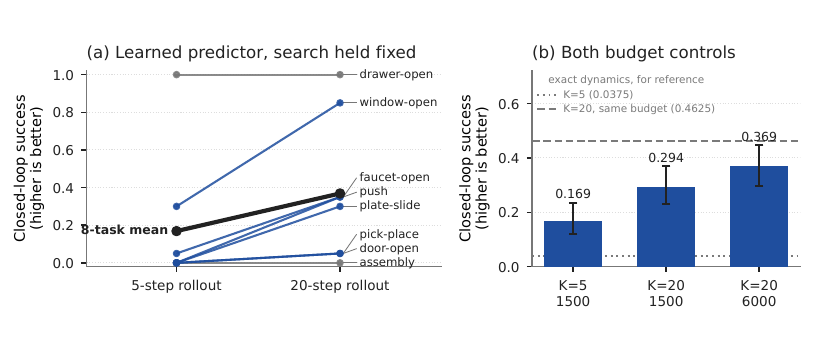}
\caption{Matching the rollout to the goal in closed loop with the learned predictor. \textbf{(a)} Every task at a five-step and a twenty-step rollout with the search held fixed. \textbf{(b)} Both budget controls, with 95\% Wilson intervals. The dashed lines are the exact-dynamics reference, shown for comparison only.}
\label{fig:closedloopK}
\end{figure}

Eight tasks are evaluated with twenty episodes each, using the V-JEPA 2.1 predictor trained with five recursive steps and demonstration targets twenty control steps ahead. The three conditions differ only in the rollout horizon and the search budget, written as candidates $\times$ rollout steps $\times$ iterations. The first rolls five steps at $100\times5\times3$. The second rolls twenty steps at $100\times20\times3$, which holds the search fixed and pays for the longer rollout. The third rolls twenty steps at $25\times20\times3$, which holds the total budget at 1500 and shrinks the search instead. Table~\ref{tab:closedloopK} reports all three and Figure~\ref{fig:closedloopK} shows them by task.

Holding the search fixed raises success from 16.9\% to 36.9\%. Holding the budget fixed raises it to 29.4\%. Both controls move in the same direction, so the gain does not come only from additional computation. Paired by episode, the longer rollout wins on 32 episodes and loses on none at fixed search, and wins on 26 and loses on 6 at equal budget. Both differences are significant in the exact McNemar test ($p=4.7\times10^{-10}$ and $p=5.4\times10^{-4}$). At fixed search no task becomes worse. At equal budget one task, drawer-open, falls from 100\% to 75\%, where the twenty-step rollout uses a quarter of the candidates.

Under exact dynamics with a state-based distance, the same eight tasks move from 3.8\% to 46.3\% at the same equal budget, and to 55.0\% at the full search. This pair is shown for reference only. The learned predictor achieves about 29\% of the exact-dynamics gain at equal budget and about 39\% at full search. Part of this gap may come from prediction errors, but the two planners also use different scores, latent cosine distance against true-state distance.

\begin{table}[htbp]
\centering
\caption{Closed-loop success with the learned predictor at a five-step and a twenty-step rollout, on eight tasks with twenty episodes each. Budgets are candidates $\times$ rollout steps $\times$ iterations. The exact-dynamics rows are measured with true simulator transitions and a state-based distance, and are shown for reference only.}
\label{tab:closedloopK}
\begin{tabular}{lrrr}
\toprule
Task & $K=5$ & $K=20$ & $K=20$ \\
 & $100\times5\times3$ & $25\times20\times3$ & $100\times20\times3$ \\
 & 1500 & 1500 & 6000 \\
\midrule
assembly & 0.00 & 0.00 & 0.00 \\
drawer-open & 1.00 & 0.75 & 1.00 \\
door-open & 0.00 & 0.00 & 0.05 \\
pick-place & 0.00 & 0.05 & 0.05 \\
plate-slide & 0.00 & 0.35 & 0.30 \\
push & 0.05 & 0.15 & 0.35 \\
faucet-open & 0.00 & 0.25 & 0.35 \\
window-open & 0.30 & 0.80 & 0.85 \\
\midrule
Mean & 0.1688 & 0.2938 & \textbf{0.3688} \\
Successes & 27/160 & 47/160 & 59/160 \\
Gains/losses against $K=5$ & --- & 26/6 & 32/0 \\
Exact McNemar $p$ & --- & $5.4\times10^{-4}$ & $4.7\times10^{-10}$ \\
\midrule
\multicolumn{4}{l}{\emph{Exact dynamics with a state-based distance, shown for reference only.}} \\
\quad Exact dynamics & 0.0375 & 0.4625 & --- \\
\bottomrule
\end{tabular}

\end{table}

\paragraph{Sixteen tasks with a single goal image.} We repeat the change on all sixteen main tasks without demonstration targets (Table~\ref{tab:x30}). The planner receives only an image of the completed task. In the main setting, this image is the final frame of an expert rollout from the same episode. A second setting uses the final frame of an expert rollout from another reset of the task, whose object positions can differ. Budgets are 100$\times$5$\times$3 for the five-step planner, 17$\times$30$\times$3 for the thirty-step planner at the same total budget, and 100$\times$30$\times$3 for the full search. Each thirty-step row is compared with the five-step row of the same setting over 320 paired episodes. With MPC and the goal image from the same episode, no task becomes worse at either budget. With pure imagination, one task becomes worse (door-close, 85\% to 35\%). With the goal image from another reset, seven tasks improve and five become worse.

\begin{table}[htbp]
\centering
\caption{Closed-loop success on the sixteen main tasks with the V-JEPA 2.1 predictor trained with five recursive steps, twenty paired episodes per task, and no subgoals. MPC observes the environment after every control step; ``none'' plans from imagined states only. Budgets are candidates $\times$ rollout steps $\times$ iterations. Gains and losses are paired episodes against the five-step row of the same setting.}
\label{tab:x30}
\begin{tabular}{llrrrr}
\toprule
Rollout $K$ & Feedback & Budget & Success & Gains/losses & Exact McNemar $p$ \\
\midrule
\multicolumn{6}{l}{\emph{Image of the completed task, same episode}} \\
5 & MPC & $100\times5\times3$ & 0.3031 & --- & --- \\
30 & MPC & $17\times30\times3$ & 0.4656 & 57/5 & $3.1\times10^{-12}$ \\
30 & MPC & $100\times30\times3$ & 0.5156 & 70/2 & $1.1\times10^{-18}$ \\
5 & none & $100\times5\times3$ & 0.2250 & --- & --- \\
30 & none & $100\times30\times3$ & 0.3594 & 58/15 & $4.1\times10^{-7}$ \\
\multicolumn{6}{l}{\emph{Image of the completed task, another reset}} \\
5 & MPC & $100\times5\times3$ & 0.2469 & --- & --- \\
30 & MPC & $100\times30\times3$ & 0.3219 & 43/19 & $3.2\times10^{-3}$ \\
\bottomrule
\end{tabular}

\end{table}

\section{Harder alternatives}
\label{app:hardneg}
The comparison between encoders returns once the alternatives are hard to beat. Appendix~\ref{app:khorizon} reports that at a matched rollout the expert percentile against random candidates collapses to zero. A percentile of zero means the expert beat all 100 alternatives in every scored window, which is perfect discrimination, and a perfect score cannot separate a good encoder from a better one. The measurement is saturated there rather than equal.

We therefore replace the random alternatives with perturbations of the expert's own actions, at a deviation $\sigma$, and repeat the measurement at a matched rollout. Smaller $\sigma$ is the harder test, because the alternatives sit closer to the expert. Table~\ref{tab:hardneg} reports the resulting expert percentile and Figure~\ref{fig:hardneg} shows it. These values are a different quantity from the percentile defined in Appendix~\ref{app:pstar}, which is measured against a clipped standard Gaussian, so the two are not compared directly.

Against these harder alternatives, V-JEPA 2.1 still ranks the expert reliably thirty steps ahead. At $\sigma=0.25$ and a thirty-step rollout the mean percentile is 6.54, below the reporting threshold of 10, whereas at $\sigma=0.1$ it is 21.64. The twenty-step and thirty-step means use different task sets, of four tasks and three, because door-close is too short for a thirty-step rollout, so they are not compared directly.

At a matched twenty-step rollout with $\sigma=0.25$ the five encoders separate again, from 13.02 for V-JEPA 2.1 to 35.43 for VideoPrism. The spread is 22.41 percentile points, whereas the same five encoders spread by 0.34 against random alternatives. The window counts are identical across all five encoders on every task, at 49, 40, and 60, so this is the same comparison on the same windows. V-JEPA 2.1 ranks first here and second under the five-step evaluation, whereas DINOv2 moves from first to fourth. Encoder comparisons should therefore be reported together with the candidate distribution used.

\begin{table}[htbp]
\centering
\caption{Expert percentile against perturbed-expert alternatives at a matched rollout. Lower is better. Panel (a) varies the perturbation $\sigma$, where smaller is harder, and the two rows are computed on different task sets. Panel (b) compares the five encoders at $K=20$ and $\sigma=0.25$ beside the same encoders measured against random Gaussian alternatives, which is a different quantity reported for context only.}
\label{tab:hardneg}
\begin{tabular}{lrrr}
\toprule
\multicolumn{4}{l}{\emph{(a) Expert percentile against perturbed-expert alternatives}} \\
\midrule
Matched rollout & $\sigma=0.1$ & $\sigma=0.25$ & $\sigma=0.5$ \\
\midrule
$K=20$ (4 tasks) & 25.00 & 12.25 & 2.46 \\
$K=30$ (3 tasks) & 21.64 & 6.54 & 0.64 \\
\midrule
\multicolumn{4}{l}{\emph{(b) The five encoders at $K=20$, $\sigma=0.25$, on three tasks}} \\
\midrule
Encoder & Perturbed-expert & Random Gaussian \\
\midrule
V-JEPA 2.1 & 13.02 & 0.00 \\
V-JEPA 2 & 20.70 & 0.00 \\
VideoMAEv2 & 28.90 & 0.02 \\
DINOv2 & 33.73 & 0.00 \\
VideoPrism & 35.43 & 0.34 \\
\midrule
Spread & \textbf{22.41} & 0.34 \\
\bottomrule
\end{tabular}

\end{table}

\begin{figure}[htbp]
\centering
\includegraphics[width=\textwidth]{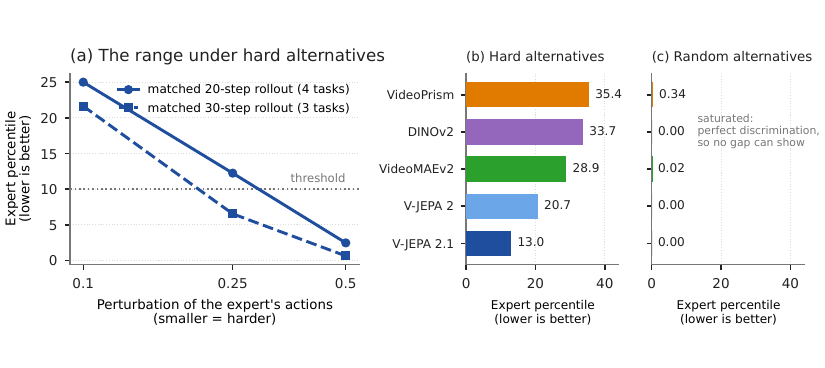}
\caption{Encoder differences return when the alternatives are hard. Lower is better in every panel. \textbf{(a)} Expert percentile against perturbed-expert alternatives by perturbation size, at two matched rollouts measured on different task sets. \textbf{(b)} The five encoders against hard alternatives. \textbf{(c)} The same five encoders where the random-alternative measurement saturates. Panels (b) and (c) use separate axes.}
\label{fig:hardneg}
\end{figure}

\section{Comparing prediction and ranking under one protocol}
\label{app:protocol}
Recorded-action identification can be measured in two ways. The model can be unrolled all the way to the goal, or the rollout can be held at five steps while the goal moves, as in expert action ranking. Comparing identification under the first protocol with ranking under the second compares a model given the goal's full lookahead against one given five steps of it, so any gap between the two curves reflects the procedures rather than two capabilities.

\begin{figure}[htbp]
\centering
\includegraphics[width=\textwidth]{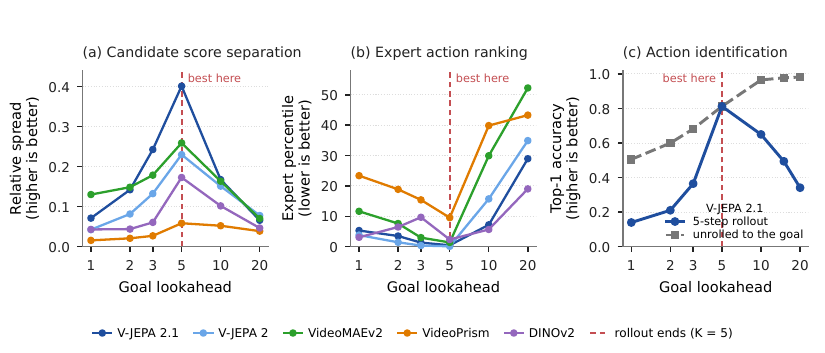}
\caption{Three measurements under one protocol, each best where the rollout ends. \textbf{(a)} Candidate score separation, where higher is better. \textbf{(b)} Expert action ranking, where lower is better. \textbf{(c)} Recorded-action identification for V-JEPA 2.1 under both protocols, where higher is better. The vertical line marks the five-step rollout.}
\label{fig:protocol}
\end{figure}

We therefore measure identification under the ranking protocol, holding the rollout at five steps and moving the goal, so that both curves use one procedure. Table~\ref{tab:protocol} reports both for V-JEPA 2.1 and Figure~\ref{fig:protocol} places them beside the other two measurements. Identification then decays in the same way that ranking does. At a goal twenty control steps ahead, identification is 0.982 when the model is unrolled to the goal but 0.342 under the five-step rollout, and the five-step curve peaks at a goal five steps ahead, where the rollout ends. The two protocols agree exactly at that point, at 0.812, where they are the same computation. For all five encoders, identification under the five-step rollout peaks at a goal five steps ahead and falls on both sides of it.

Under this shared protocol, prediction and ranking behave alike. Both are best when the goal sits at the end of the rollout and both fall away on either side of it, which is the same pattern that candidate score separation shows in Table~\ref{tab:dispersion}. Three measurements on five encoders therefore locate the same peak, at the end of the rollout.

\begin{table}[htbp]
\centering
\caption{Recorded-action identification for V-JEPA 2.1 under both protocols. The first protocol unrolls the model to the goal; the second holds the rollout at five steps and moves the goal. The two agree exactly at a goal five steps ahead, where they are the same computation.}
\label{tab:protocol}
\begin{tabular}{lrrrrrrr}
\toprule
Protocol & $L=1$ & $L=2$ & $L=3$ & $L=5$ & $L=10$ & $L=15$ & $L=20$ \\
\midrule
Unrolled to the goal & 0.506 & 0.600 & 0.682 & 0.812 & 0.965 & 0.978 & 0.982 \\
Five-step rollout & 0.140 & 0.211 & 0.365 & \textbf{0.812} & 0.650 & 0.494 & 0.342 \\
\bottomrule
\end{tabular}

\end{table}

\section{Predictor capacity, training horizon and search budget}
\label{app:ladder}
The capacity comparison evaluates the 3.75\,M-parameter baseline and five larger predictors, up to 305.5\,M parameters. Aggregate $P^*$ remains ten on matched diagnostic windows with the five-step rollout. At a matched rollout, the test against random alternatives saturates for all predictors (Appendix~\ref{app:hardneg}), so the capacity comparison is reported at the five-step rollout, where it can separate them. Capacity improves prediction-related metrics, and the largest model reaches validation loss within 1.5\% of the best capacity setting. The step from the baseline to the next capacity also adds training data.

\begin{figure}[htbp]
\centering
\includegraphics[width=0.65\textwidth]{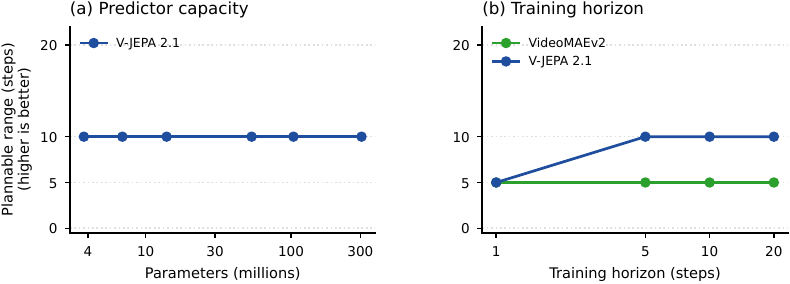}
\caption{Plannable range with a five-step rollout under changes to \textbf{(a)} predictor capacity, including the 3.75\,M-parameter baseline, and \textbf{(b)} number of recursive training steps. Higher is better; both panels use four-task aggregate curves.}
\label{fig:ladder}
\end{figure}

The training-horizon comparison uses one, five, ten, and twenty recursive training steps. V-JEPA 2.1's range is five for the predictor trained with one step and ten for the others; the evaluation rollout remains five steps. We repeat the comparison on five tasks with a thirty-step rollout that reaches targets thirty steps ahead. For both V-JEPA 2.1 and VideoMAEv2, the predictor trained with five recursive steps already ranks the expert action within the reporting threshold, and training with more recursive steps improves the percentile by at most 3.2 points. Once the rollout reaches the target, the training horizon therefore has little further effect: the rollout used during planning, rather than the one used during training, determines how far ahead the model can guide action selection. Figure~\ref{fig:ladder} reports the measured range for the displayed settings.

All reported planners use cosine distance, which performed at least as well as learned goal-scoring heads trained on demonstrations in closed-loop control.

The search-budget comparison increases the CEM candidate count and elite count tenfold, preserving elite fraction, iteration count, rollout horizon, and objective. Both the baseline and the larger budget have 0/60 successes on the three tested failure tasks. A larger search therefore does not recover these failures, which is consistent with the limit lying in the scoring configuration rather than in the search.

\section{Latent distance and expert-action ranking}
\label{app:weakest}
\begin{figure}[htbp]
\centering
\includegraphics[width=0.65\textwidth]{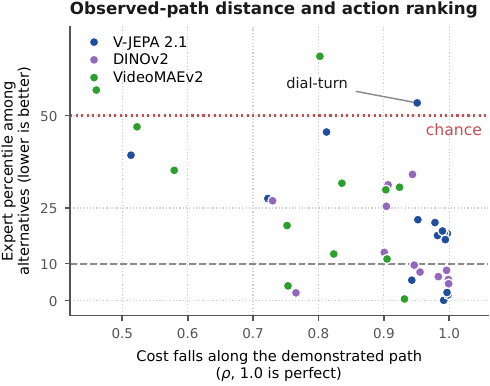}
\caption{Target distance along observed demonstrations and expert-action ranking through learned rollouts. The two measurements use observed and predicted states, respectively.}
\label{fig:why}
\end{figure}

Distance to a goal can decrease along an observed demonstration while failing to favour the expert's short action sequence over random candidates scored through a learned rollout (Figure~\ref{fig:why}). The first measurement uses observed features and the second predicted features. The exact-dynamics experiment separates the two: distant-goal failures also occur without a learned predictor (Appendix~\ref{app:exact}).

Cosine and squared Euclidean scoring are compared over 421 matched goal settings. Four of five encoders have the same range under these two scores, and squared Euclidean scoring shortens the fifth. This is consistent with both criteria comparing a short trajectory with a target state, without estimating whether that target is reachable. Objectives that estimate temporal distance or task value are a natural extension.

\section{Predicting where planning works}
\label{app:screen}
\begin{figure}[htbp]
\centering
\includegraphics[width=0.75\textwidth]{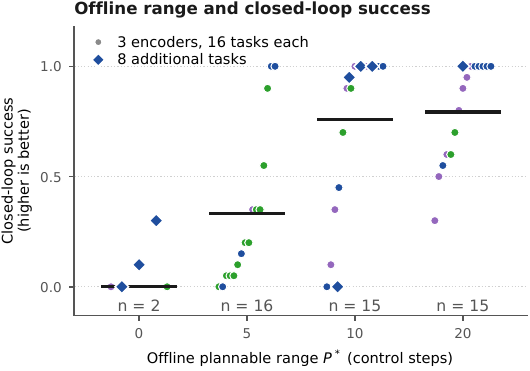}
\caption{Closed-loop success grouped by offline plannable range. Circles denote encoder--task pairs, diamonds denote eight additional tasks, and horizontal marks show group means.}
\label{fig:screen}
\end{figure}

The plannable range can be used to predict which tasks a planner will solve. For eight additional tasks, the offline range predicts closed-loop success under a simple rule: success of at least 95\% for a positive range and at most 30\% for a zero range. Seven of the eight predictions are correct. Across the thirteen training tasks and three encoders, encoder--task pairs with a positive range also reach higher closed-loop success (Figure~\ref{fig:screen}). The offline range further indicated that extending the rollout to reach the target would improve control, which Appendix~\ref{app:closedloopK} confirms with the learned predictor. The range compares configurations that share a scoring function; it is not designed to compare different scoring functions.

\section{Statistical tests}
\label{app:stats}

Success plots show Wilson 95\% binomial intervals over episode counts. With equal episode counts, pooled success also equals the mean task success. These intervals describe the evaluated episodes. McNemar tests compare binary episode outcomes paired by task and evaluation index. Corresponding episodes are reset with seeds $10000+e$ and use the same scripted-expert inclusion filter.

The three-regime comparison uses an exact two-sided sign test over twelve matched encoder--task conditions, excluding tied differences. To supplement episode-level selection tests, we resample whole tasks with replacement, retaining all episode pairs within each task, and compute 10\,000 bootstrap mean differences with random seed zero. Table~\ref{tab:uncertainty} gives percentile intervals. These describe variation over the observed task sets.

\begin{table}[htbp]
\centering
\caption{Paired success differences and 95\% task-bootstrap intervals. Positive values favour the second named system.}
\label{tab:uncertainty}
\begin{tabular}{lrr}
\toprule
Comparison & Difference & 95\% interval \\
\midrule
VLA to VLA--WM (16 tasks) & 0.122 & [0.038, 0.219] \\
WM to simulator critic (7 tasks) & 0.057 & [-0.007, 0.121] \\
VLA to VLA--WM (4 near tasks) & 0.237 & [0.075, 0.412] \\
\bottomrule
\end{tabular}

\end{table}

\section{The exact-dynamics reference}
\label{app:exact}
The simulator control saves the full environment state, branches each candidate action sequence, and restores the saved state before the next candidate and before real execution. State restoration is checked by replaying actions from the same saved state. No learned encoder or predictor is used to evaluate candidate outcomes in this control.

The state vector contains hand position, gripper opening, and object position. The score is the minimum unweighted Euclidean distance to the target state over the simulated trajectory. The subgoal is selected by locating the nearest demonstrated state under the same distance and advancing by $L$ control steps. This state-based distance differs from the learned world model's latent cosine score.

The standalone simulator planner uses CEM with 100 candidates, ten elites, and three iterations per decision, with a five-control-step horizon and warm starts from the previous solution. The first two environment actions are executed before replanning. Table~\ref{tab:exact} changes goal lookahead from five to twenty while keeping this search horizon fixed. The shorter goal gives 165 paired gains and no losses across the sixteen tasks. Exact prediction therefore does not remove the distant-goal failure when the rollout stays short, and Appendix~\ref{app:closedloopK} shows that extending the rollout reduces it. Figure~\ref{fig:teaser}(b) compares these results with the learned world model.

\begin{table}[htbp]
\centering
\caption{Closed-loop success with simulator dynamics and a fixed five-step rollout. Only the target lookahead changes. The comparison removes learned prediction error but retains the mismatch between rollout length and target distance in the second row.}
\label{tab:exact}
\begin{tabular}{lrrr}
\toprule
Target lookahead & Training & Held out & All sixteen \\
\midrule
5 control steps & \exactFiveTrained{} & \exactFiveHeldout{} & \exactFive{} \\
20 control steps & \exactTwentyTrained{} & \exactTwentyHeldout{} & \exactTwenty{} \\
\bottomrule
\end{tabular}
\end{table}

The simulator-based VLA critic evaluates VLA chunks generated with the same settings as VLA--WM, using their first ten environment actions, and executes the first block of the lowest-scoring chunk.

\section{Acting without observations}
\label{app:blind}
The blind-step comparison withholds new observations for one, two, five, ten, or twenty-five control steps. The world model continues to predict states and replan in imagination between observation updates. The VLA baseline executes the corresponding portion of its previously predicted fifty-action chunk. At two environment actions per control step, twenty-five control steps exhaust the VLA chunk; no actions are repeated to extend it. Both systems use the same observation-update interval on the horizontal axis. Near variants and new tasks are shown separately. On near variants, the world model keeps success above 90\% for up to ten control steps without new observations, close to the offline plannable range, whereas the VLA remains below 30\%. Without observations for twenty-five control steps, both systems fail.

\begin{figure}[htbp]
\centering
\includegraphics[width=0.7\textwidth]{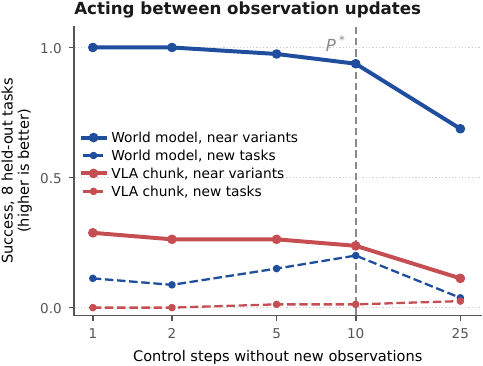}
\caption{Success versus control steps without new observations on four near variants and four new tasks. The world model replans from predicted states, whereas the VLA executes its predicted chunk. The vertical line marks the aggregate offline range.}
\label{fig:blind}
\end{figure}

\begin{figure}[htbp]
\centering
\includegraphics[width=0.5\textwidth]{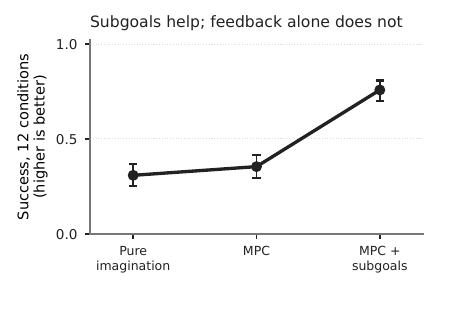}
\caption{Pure imagination, MPC, and MPC with subgoals five steps ahead over twelve encoder--task conditions (three encoders, four diagnostic tasks, five-step rollout). Error bars show 95\% Wilson intervals.}
\label{fig:regimes}
\end{figure}

An earlier comparison of pure imagination, MPC, and MPC with subgoals used twelve conditions: three encoders (V-JEPA 2.1, VideoMAEv2, and DINOv2) on the four diagnostic tasks, all with a five-step rollout (Figure~\ref{fig:regimes}). Without subgoals, the target is the final observation of an expert rollout from a different reset of the task. Success is 30.8\% with pure imagination and 35.4\% with MPC, a difference that is not significant in a sign test over the twelve conditions ($p=0.69$). Subgoals five steps ahead raise success to 75.8\% ($p=0.012$). An untrained predictor given the same subgoals reaches only 0.8\%. Section~\ref{sec:pure} repeats the comparison with V-JEPA 2.1 on all sixteen tasks.

\section{VLA fine-tuning, selection and the failure profile}
\label{app:profile}
\subsection{Standalone VLA and world-model performance}

We first compare a fine-tuned $\pi_0$ VLA with the standalone V-JEPA 2.1 planner, which motivates their combination. The VLA receives RGB, a task instruction, and a seven-dimensional simulator-state vector. The world-model planner additionally receives demonstration subgoals taken from an expert rollout of the same episode, so the comparison places the two systems in context rather than ranking them.

\begin{figure}[htbp]
\centering
\includegraphics[width=\textwidth]{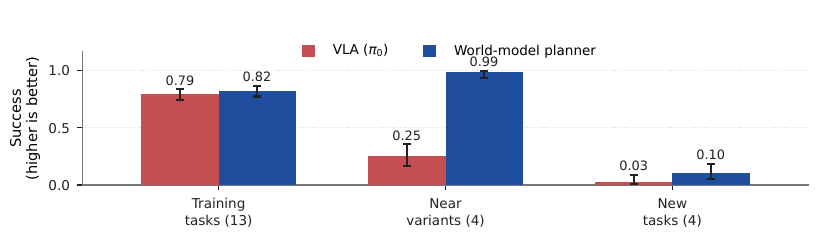}
\caption{Standalone VLA and world-model success on thirteen training tasks, four near variants, and four tasks with new objects or motion patterns. The world-model planner receives demonstration subgoals taken from an expert rollout of the same episode; error bars show 95\% Wilson intervals.}
\label{fig:profile}
\end{figure}

With one fine-tuned VLA and one world-model planning configuration, their success rates are 79.2\% and 82.3\% on the thirteen training tasks, 25.0\% and 98.8\% on four near variants of trained object families, and 2.5\% and 10.0\% on four tasks involving new objects or motion patterns (Figure~\ref{fig:profile}). The two perform similarly on the training tasks, while the world-model planner, which also receives subgoals, reaches much higher success on the near variants. Both remain near the floor on the new tasks. This profile motivates using the VLA to propose task-conditioned actions and the world model to evaluate their predicted outcomes.

\begin{table}[htbp]
\centering
\caption{Success on held-out tasks. The VLA--WM and standalone world-model methods use demonstration subgoals from expert rollouts of the same episodes. Their candidate-generation procedures differ, so the table compares observed capabilities rather than providing a controlled ranking.}
\label{tab:heldout_pairing}
\begin{tabular}{lcc}
\toprule
Method & Near variants & New tasks \\
\midrule
VLA & 25.0\% & 2.5\% \\
VLA + world-model selection & 48.7\% & 6.2\% \\
Standalone world-model planner & 98.8\% & 10.0\% \\
\bottomrule
\end{tabular}
\end{table}

% --- Held back for the rebuttal (not shown in the PDF). ---
% \subsection{The source of the subgoal}
% \label{app:subgoalsource}
% The selection gain in Section~\ref{sec:method} uses a demonstration from the same environment reset as the evaluated episode. We also evaluate two subgoal sources that do not require this alignment, on the same sixteen tasks with twenty paired episodes per task. With one demonstration from a different reset, selection changes success relative to the VLA baseline by $-0.066$ ($p=0.0046$). With the closest demonstration retrieved from a library of five, the change is $+0.009$ ($p=0.766$). A misaligned subgoal can therefore direct the world model towards the wrong state, and obtaining an aligned subgoal without an expert rollout remains an open problem.

\subsection{Fine-tuning and inputs}
We adapt the released $\pi_0$ base checkpoint using the LeRobot implementation. The dataset contains 939 training and 237 validation episodes, corresponding to 124\,050 and 31\,350 action-chunk starting points. The split is by episode, with every fifth episode assigned to validation. Inputs are a $224\times224$ RGB image, a task instruction tokenised with the PaliGemma tokenizer, and the seven-dimensional current state. Targets contain fifty four-dimensional environment actions. The model pads state and action widths internally to 32; the loss uses the four actual action dimensions. Episode-end chunks repeat the last available action and retain padding indicators.

LoRA is applied to attention query, key, value, and output projections, with rank 32, scale 64, and zero dropout. Training uses batch size four, one epoch, seed zero, and bfloat16 arithmetic on an NVIDIA L40S. AdamW uses learning rate $2.5\times10^{-5}$, decay to $2.5\times10^{-6}$, 1000 warm-up steps, momentum parameters $(0.9,0.95)$, weight decay 0.01, and gradient clipping at one. Validation is evaluated every 2000 updates over 200 batches. The run completes 31\,012 updates; the selected adapter is the lowest-validation-loss checkpoint at update 28\,000.

State and action means and standard deviations are computed from training data and saved with the adapter. Inference uses these same statistics and maps actions back to environment units. The dataset checks verify disjoint episode splits, chunk alignment, padding, image range, actual action width, language inputs, and normalisation round trips.

\subsection{Candidate selection and execution}
The main selection condition draws eight chunks with independent flow-matching initial noise at temperature 1.0. Each candidate contributes its first ten actions, grouped into five two-action blocks. The learned world model evaluates Equation~\ref{eq:score}, and the controller executes the first block of the best-ranked candidate. The VLA-only baseline queries the policy once per environment step, so a two-action control interval consists of two independent draws taken at the same observation, and both are executed. Both conditions therefore execute two actions per control interval and receive new observations at the same control rate; they differ in where the two actions come from, independent draws for the baseline and the chosen chunk for selection. We query once per environment step because the adapter is supervised at that rate, one action per environment step; holding one action for the whole interval would not match how the adapter was trained.

\subsection{The baseline executes the same number of actions}
A second baseline confirms that the selection gain does not come from how the baseline executes actions. Because the two actions of the baseline's block are drawn independently rather than taken from a single chunk, we also report a baseline that executes one chunk's first two actions, which is the same policy at the same observation rate with no world model in the loop. Over sixteen tasks and twenty episodes it reaches 63.8\%, against 65.0\% for the per-step baseline, and the two are indistinguishable in the episode-paired test, at 15 gains and 19 losses with $p=0.608$. Selection improves on this matched baseline by 13.4 percentage points, at 45 gains and 2 losses with $p=1.6\times10^{-11}$, slightly more than the 12.2 points it improves on the per-step baseline, and it wins on nine tasks, loses on none, and ties on seven.

Both baselines receive new observations at the same control rate as selection, and neither repeats an action to fill the interval. The three conditions therefore differ only in where the two executed actions come from: two independent draws for the per-step baseline, one chunk's first two actions for the matched baseline, and the chosen chunk for selection.

Demonstration-goal conditions use an expert rollout from the same environment reset as the evaluated episode. Goal observations are sampled every two environment actions. The world model localises by highest cosine similarity to the demonstration and advances five points, with the index capped at the final point. It does not require a separate distance threshold to advance. Task-goal conditions instead use a final goal from a separate reset. Evaluation ends at the environment success condition or the 200-environment-step limit; episodes whose scripted expert fails the inclusion check are omitted consistently across conditions.

The seven-task subset contains the tasks where VLA top-1 success is below 95\%: assembly, dial-turn, handle-pull, pick-place, push, reach, and window-open. It contains four training tasks and the three main held-out tasks. The full sixteen-task result is always reported alongside it.

The learned critic recovers 82.2\% of the improvement achieved by simulator-based scoring on the seven-task subset, and the difference between the two critics is not significant in the episode-paired test ($p=0.134$). The simulator critic uses true-state distance and state-based subgoal localisation, whereas the learned critic uses latent cosine distance. Per-task outcomes are shown in Figure~\ref{fig:pertask} in Appendix~\ref{app:pertask}.

\begin{figure}[htbp]
\centering
\includegraphics[width=\textwidth]{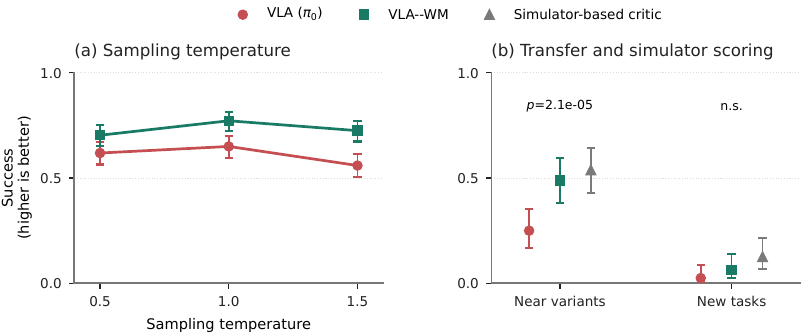}
\caption{Additional VLA--WM evaluations. \textbf{(a)} Sampling-temperature sweep with a matched VLA baseline on sixteen tasks. \textbf{(b)} Learned and simulator-based critics on near variants and new tasks. Error bars show 95\% Wilson intervals; annotations use paired episode outcomes.}
\label{fig:robustness}
\end{figure}

Candidate counts one, eight, sixteen, and thirty-two are compared on this subset. Temperature comparisons use 0.5, 1.0, and 1.5, each with its own matched VLA baseline. Selection improves success at each tested temperature relative to its matched VLA baseline (Figure~\ref{fig:robustness}(a)).

\section{Per-task results}
\label{app:pertask}
Table~\ref{tab:pertask} gives the main VLA, VLA--WM, and simulator-critic results as successes over evaluated episodes. Held-out tasks are marked in the table. The simulator critic shares the VLA proposal protocol but changes both outcome prediction and the scoring geometry, as described in Appendix~\ref{app:exact}. Additional temperature results appear in Figure~\ref{fig:robustness}(a), and task-group results with simulator-based scoring in Figure~\ref{fig:robustness}(b).

\begin{table}[htbp]
\centering
\caption{Main evaluation counts at eight candidates for selection conditions. The VLA baseline executes one policy sample per environment step (Appendix~\ref{app:profile}).}
\label{tab:pertask}
\begin{tabular}{llrrr}
\toprule
Task & Split & VLA & VLA--WM & Simulator \\
\midrule
assembly & train & 5/20 & 15/20 & 14/20 \\
button press topdown & train & 20/20 & 20/20 & 20/20 \\
coffee button & train & 20/20 & 20/20 & 20/20 \\
dial turn & train & 10/20 & 19/20 & 20/20 \\
door close & train & 20/20 & 20/20 & 20/20 \\
door open & train & 19/20 & 20/20 & 20/20 \\
drawer close & train & 20/20 & 20/20 & 20/20 \\
drawer open & train & 20/20 & 20/20 & 20/20 \\
faucet open & train & 20/20 & 20/20 & 20/20 \\
handle press & train & 20/20 & 20/20 & 20/20 \\
handle pull & held out & 0/20 & 0/20 & 0/20 \\
pick place & train & 4/20 & 12/20 & 16/20 \\
plate slide & train & 19/20 & 20/20 & 20/20 \\
push & held out & 0/20 & 2/20 & 1/20 \\
reach & train & 9/20 & 17/20 & 20/20 \\
window open & held out & 2/20 & 2/20 & 4/20 \\
\bottomrule
\end{tabular}

\end{table}

\begin{figure}[htbp]
\centering
\includegraphics[width=\textwidth]{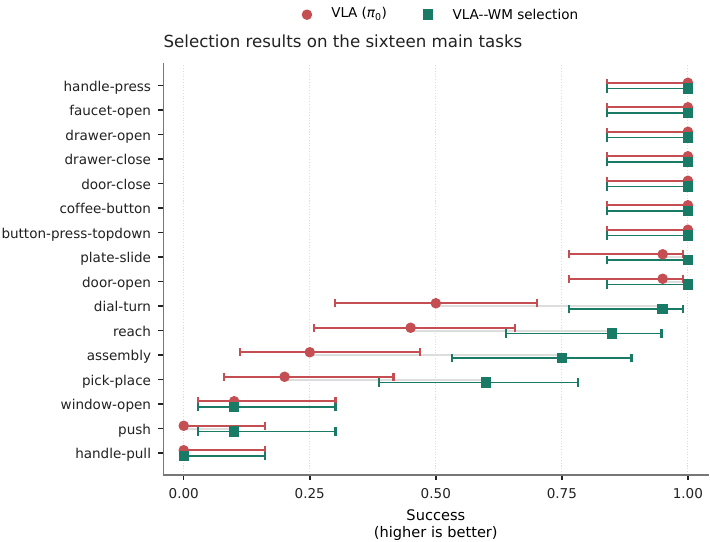}
\caption{Per-task success of VLA and VLA--WM across the sixteen main tasks. Each task uses twenty episodes; horizontal error bars show 95\% Wilson intervals.}
\label{fig:pertask}
\end{figure}

\section{Scope of the results}
\label{app:limits}
\paragraph{Simulation and embodiment.} Closed-loop results use Meta-World with one robot arm. This choice allows the learned predictor to be replaced by exact simulator dynamics with everything else held fixed, which is how we separate prediction error from the scoring configuration. Offline real-robot data from 2\,958 WidowX episodes show the same drop in action ranking with target distance, despite a different embodiment, action space and control rate.

\paragraph{Tasks and seeds.} The plannable range is measured on four articulated-object tasks, which provide expert trajectories long enough to test lookaheads of twenty and thirty steps. Closed-loop evaluations cover sixteen main tasks and eight additional tasks, with paired tests at the episode level. Most comparisons use one training seed; where additional seeds are available, they agree closely (Table~\ref{tab:dispersion}).

\paragraph{Lower bounds.} When the rollout reaches the target, the range reaches the largest lookahead tested and is therefore reported as a lower bound. Task length limits how far this can be tested, because few tasks provide enough expert steps beyond thirty.

\paragraph{Rollout length.} We test rollouts of up to thirty control steps. The results support extending the rollout until it reaches the target, not using the longest possible rollout. A longer rollout needs more computation, and at the same total budget the planner must evaluate fewer candidates. The learned model also gains less than exact dynamics: with a twenty-step rollout, it obtains 29--39\% of the exact-dynamics gain, although the two planners also use different scores (Appendix~\ref{app:closedloopK}). With pure imagination, one task becomes worse with a thirty-step rollout (Appendix~\ref{app:closedloopK}). Offline, ranking remains reliable when the rollout extends past the target: with a twenty-step rollout, the expert percentile stays below 2.5 for every target up to ten steps ahead (Table~\ref{tab:kcurve}), because the score keeps the closest predicted state.

\paragraph{Subgoals.} The control gains in Sections~\ref{sec:pure} and~\ref{sec:method} use subgoals from an expert rollout of the same episode. This setting isolates whether the world model can use a nearby target, which is the question these sections ask. Generating such subgoals without an expert rollout is a natural next step.

\paragraph{Diagnostic and optimiser.} The plannable range is a goal-ranking diagnostic rather than an optimiser-independent planning bound. Its dependence on the rollout length was measured offline and then confirmed in closed loop with the learned predictor (Appendix~\ref{app:closedloopK}). Its dependence on the reporting threshold and on the candidate distribution is reported in Table~\ref{tab:thresholds} and Appendix~\ref{app:hardneg}.

% --- Held back for the rebuttal (not shown in the PDF). ---
% \paragraph{Computation.} In our implementation, the combined VLA and world-model system takes longer per control step than either component alone, because it encodes and scores several proposals at each decision.

\end{document}